\documentclass[lettersize,journal]{IEEEtran}
\usepackage{amsmath,amsfonts}
\usepackage{algorithmic}
\usepackage{algorithm}
\usepackage{array}
\usepackage[caption=false,font=normalsize,labelfont=sf,textfont=sf]{subfig}
\usepackage{textcomp}
\usepackage{stfloats}
\usepackage{url}
\usepackage{verbatim}
\usepackage{graphicx}
\usepackage{cite}
\usepackage{mathrsfs}
\usepackage{multirow}
\usepackage{makecell}
\usepackage{pifont} 
\usepackage{xcolor}
\usepackage[pagebackref,breaklinks,colorlinks]{hyperref}

\usepackage{xfrac} 
\usepackage[normalem]{ulem}

\begin{document}

\title{Disentanglement-Based Equivariant Learning for Compositional VQA}

\author{IEEE Publication Technology,~\IEEEmembership{Staff,~IEEE,}
\thanks{This paper was produced by the IEEE Publication Technology Group. They are in Piscataway, NJ.}
\thanks{Manuscript received April 19, 2021; revised August 16, 2021.}}

\author{Zhou Du,
        Zhaoquan Yuan,
        Xiao Wu,
        and~Changsheng Xu,~\IEEEmembership{Fellow,~IEEE}
\thanks{
Manuscript received December 24, 2024; revised February 21, 2025; accepted February 21, 2025. Date of publication ***; date of current version February 23, 2025.
This work was supported in part by the National Natural Science Foundation of China (Grant No. 61802053, 62372387, 62036012, U23A20387), Natural Science Foundation of Sichuan Province (Grant No. 2024NSFSC0508), the Fund of National Laboratory on Adaptive Optics, China, (Grant No. FNLAO-24-ZD-O02), Sichuan Science and Technology Program (Grant No. 2024YFHZ0029), and Key R\&D Program of Guangxi Zhuang Autonomous Region, China (Grant No. AB22080038, AB22080039).
The associate editor coordinating the review of this manuscript and approving it for publication was Prof.Wei Tsang Ooi. \textit{(Corresponding author: Zhaoquan Yuan.)}}
\thanks{
{Zhou Du, Zhaoquan Yuan, and Xiao Wu are with the School
of Computing and Artificial Intelligence, Southwest Jiaotong University,
Chengdu 611730, China, and also with Engineering Research Center of Sustainable Urban Intelligence Transportation, Ministry of Education, China (email: zhoudu@my.swjtu.edu.cn; zqyuan@swjtu.edu.cn; wuxiaohk@gmail.com).

Changsheng Xu is with the State Key Laboratory of Multimodal Artificial
Intelligence Systems (MAIS), Institute of Automation, Chinese Academy of
Sciences, Beijing 100190, China, and with the School of Artificial Intelligence, University of Chinese Academy of Sciences, Beijing 101408, China,
and also with Peng Cheng Laboratory, ShenZhen 518055, China (e-mail:
csxu@nlpr.ia.ac.cn).
}
}
}

\markboth{IEEE TRANSACTIONS ON MULTIMEDIA}%
{Shell \MakeLowercase{\textit{et al.}}: A Sample Article Using IEEEtran.cls for IEEE Journals}

\maketitle

\begin{abstract}
Compositional visual question answering (VQA) represents a challenging yet fundamental task that requires models to comprehend novel combinations of previously learned concepts.
The current methods often overlook the disentanglement of underlying concepts and are restricted in terms of their ability to effectively capture the compositional variation mechanism. 
Moreover, the state-of-the-art techniques depend on additional clues for training, which is not feasible in real-world VQA scenarios.
To address these issues, in this paper, we introduce a novel \textbf{D}isentanglement-based \textbf{E}quiv\textbf{A}riant \textbf{L}earning (\textbf{DEAL}) framework for compositional VQA, which is guided exclusively by ground-truth answers. In DEAL, we employ causality-inspired interventions to disentangle concepts derived from visual and textual inputs within a re-encoding framework.
Based on the principle of equivariance, we subsequently perform a compositional transformation on the inference input and impose the equivariant constraint on the output to augment the compositional reasoning capacity of the model.
Comprehensive experiments conducted on the benchmark CLEVR-CoGenT and GQA-SGL datasets validate the superiority of our proposed DEAL approach over the existing state-of-the-art methods for compositional VQA tasks in both visual and linguistic generalization settings.
\end{abstract}


\section{Introduction}
\IEEEPARstart{V}{isual} question answering (VQA)~\cite{Antol_2015_ICCV_VQA} is a fundamental task in the multimodal research community, where the aim is to predict target answers from provided contextual input images and questions. Significant advancements have been observed in the VQA field, especially in the context of vision-language models (VLMs)~\cite{li2022blip,wang2023image}. However, the existing methods often overlook compositional generalization capabilities, leading to reduced performance when addressing data comprising unseen visual or textual concept compositions. In contrast, humans are capable of effortlessly recognizing or generating novel combinations of familiar elementary concepts~\cite{fodor1988connectionism,bahdanau2018systematic}. This compositional generalization ability is deemed one of the essential skills that are lacking in the current artificial intelligence systems~\cite{bahdanau2019closure,ruis2020benchmark}.

\begin{figure}[t]
	\centering
	\includegraphics[width=\linewidth]{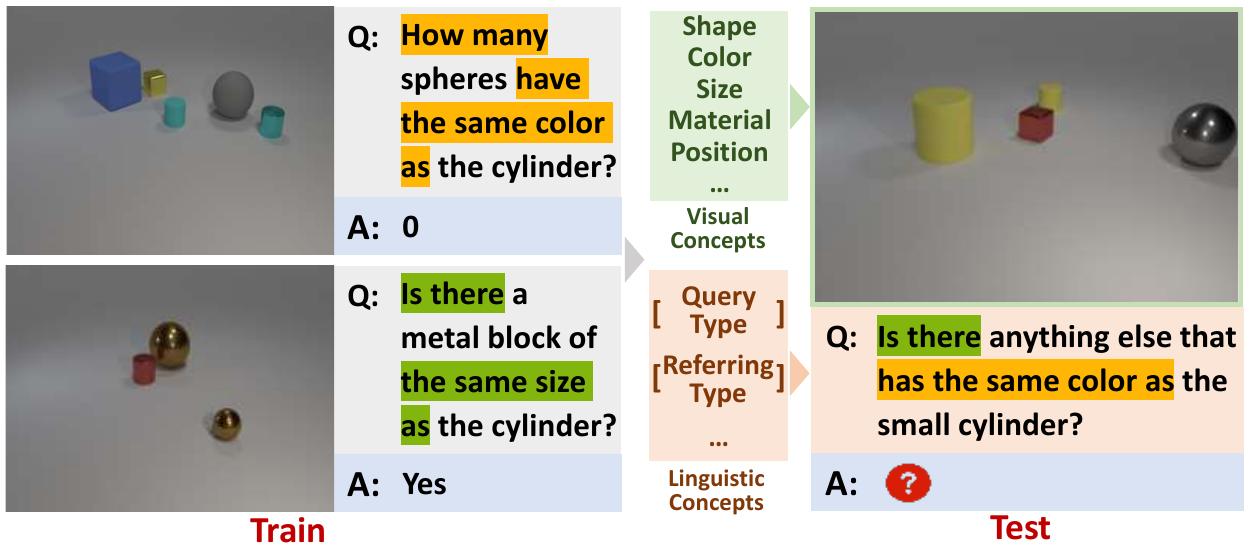}
	\caption{Illustration of the compositional VQA process with visual and linguistic generalization. The input image-question pairs can be seen as compositions of various linguistic and visual concepts. The model conducts reasoning on test inputs that contain compositions not encountered during training.}
    \label{fig_comVQA}
\end{figure}

In this paper, we focus on the task of compositional VQA.
In this task, a VQA model is trained on $\left \langle \boldsymbol{v},\boldsymbol{q},\boldsymbol{a} \right \rangle$ triplets
that encompass certain visual and linguistic concepts and is subsequently tested on image-question pairs containing unknown compositions of seen concepts. As the case depicted in Fig.~\ref{fig_comVQA} shows, the concepts “\emph{red (color)}”, “\emph{cube (shape)}”, “\emph{yes/no query (query type)}”, and “\emph{color referring (referring type)}” exist in the training samples on the left, but compositions such as $\langle$“\emph{red}”,“\emph{cube}”$\rangle$ and $\langle$“\emph{yes/no query}”,“\emph{color referring}”$\rangle$ on the right, which are novel to the model, are used for testing. 
Compositional VQA presents a critical and challenging problem with respect to robust machine reasoning.

To enhance the systematicity of VQA models, a variety of research has emerged in the field of compositional VQA. 
Some studies adaptively adjust the model's reasoning process according to input human-prior knowledge.
Among these, a few methods~\cite{bahdanau2019closure,d2021howmodular,chen2021mmn,yamada2022tmn,liu2024detection} are inspired by Neural Module Networks (NMNs)~\cite{andreas2016nmn}, which decompose the question into subtasks and address each subtask by a specific module.
Besides, approaches~\cite{cao2021lrcapsule,yi2018nsvqa,saqur2020mgn,dahlgren2023compositionalgeneralization} conduct adaptive reasoning process based on structured information of the input.
Moreover, holistic models~\cite{kamath2021mdetr,li2023exploring} have achieved success in generalization by employing extensively pre-trained models.

Nevertheless, the following limitations persist.
1) The conventional training paradigm for VQA models~\cite{yamada2022tmn,kamath2021mdetr,li2023exploring,liu2024detection} typically captures spurious correlations within concept representations, which can mislead the reasoning process.
For example, if “\emph{yellow}” and “\emph{sphere}” frequently co-occur during training, the employed model may develop an incorrect associative bias between spherical objects and the color yellow, even though there is no causal relationship between these attributes.
2) The learning procedures of~\cite{andreas2016nmn, bahdanau2019closure,cao2021lrcapsule,yamada2022tmn} overlook the compositional changes during training, which implies that their models are trained by restricted compositions derived from the provided samples without novel concept variations.
3) Most SOTA approaches depend on additional aids for training and inference purposes, such as extra guidance signals~\cite{bahdanau2019closure,d2021howmodular, yamada2022tmn, yi2018nsvqa} or pre-training knowledge~\cite{kamath2021mdetr,li2023exploring}. However, in real-world VQA scenarios, providing enough information to encompass all multimodal knowledge and reasoning frameworks is infeasible.

To overcome the limitations of the previous methods and address the challenges posed by compositional VQA, we examine the problem from the standpoint of concept learning and reconsider the following research questions.
1) What characteristics should the composable concepts that make up VQA samples possess?
2) What is the mechanism of compositional variation, and how does it influence the correctness of reasoning?
In response to the first question, we argue that composable concepts should be disentangled factors of variations. Each concept is sensitive to a single underlying generative factor while remaining relatively invariant to changes in other factors~\cite{higgins2018towards}.
Regarding the second question, the shifts between VQA samples can be considered changes in certain concepts within compositions.
We posit that when the input composition (\emph{e.g.}, $\left \langle c_1, c_2, c_3 \right \rangle$) is transformed to another composition (\emph{e.g.}, $\left \langle c_1, c_4, c_3 \right \rangle$), the corresponding output answer should equivariantly change. 
More specifically, to address the above two issues, an optimal model should first disentangle individual concepts (\emph{e.g.}, “\emph{red}”, “\emph{yellow}”, and “\emph{cube}” in Fig.~\ref{fig_comVQA}). Secondly the model is supposed to incorporate variations (\emph{e.g.}, $\langle$“\emph{red}”, “\emph{cube}”$\rangle$ to $\langle$“\emph{yellow}”, “\emph{cube}”$\rangle$) to encounter novel compositions during training and capture the equivariance between the input variations and the corresponding shifts in proper answers.

Inspired by the aforementioned motivations,
we propose a novel \textbf{D}isentanglement-based \textbf{E}quiv\textbf{A}riant \textbf{L}earning (\textbf{DEAL}) framework for compositional VQA.
In this framework, we introduce unique designs that address the aforementioned research questions and validly integrate them into a cohesive whole.
Grounded in causal intervention, which is a strategy for eliminating connections among latent factors~\cite{pearl2009causal}, we first propose an innovative concept disentanglement method.
In particular, we intervene in all encoded concept features except for a specified feature and maintain the invariance of the unchanged concept throughout the re-encoding process to eliminate noncausal correlations.
To enable the model to reason about novel compositions during training, we subsequently devise a new learning approach based on the principle of equivariance~\cite{cohen2016equivariantcnn}.
Specifically, feature-level variations are implemented as the alterations of concepts. To adhere to equivariance, the reasoning outcomes are then constrained to correspondingly transform along with the input alterations, enabling the model to generate correct answers for unseen transformed compositions.

Notably, DEAL addresses the limitations of the existing work. It achieves concept disentanglement and compositional variations during training. In addition, the ground-truth answers constitute the sole training annotations required in DEAL, which can enhance the applicability of the model across more VQA scenarios. The key contributions of our work can be summarized as follows.
\begin{itemize}
\item We propose a novel causal intervention-based disentanglement method to learn disentangled concept representations in a self-supervised manner.
In this method, a re-encoding framework is distinctively devised to guarantee the independence of each concept after conducting a unique intervention operation, which is well-suited for VQA encoding process.
\item We propose an innovative equivariant learning framework for compositional VQA, which uses a specific transformation approach and fulfills an output constraint to facilitate the model in reasoning about novel compositions. To the best of our knowledge, DEAL is the first method to address compositional generalization from the view of equivariance.
\item Extensive experiments verify the effectiveness and superiority of the proposed concept disentanglement and equivariant learning approaches. The results obtained on CLEVR-CoGenT and GQA-SGL illustrate that DEAL outperforms the current SOTA method by $1.5$\% and $2.4$\% under visual and linguistic generalization settings, respectively.
\end{itemize}

\section{Related Work}

\subsection{Visual Question Answering}
The generic VQA task is a focal point in the field of multimodal reasoning, and recent research has achieved notable developments. Traditional neural network models~\cite{ben2017mutan,ma2021joint} typically encode two modalities individually and then obtain a joint embedding. Attention-based approaches~\cite{shih2016look, mao2019neuro,liu2020visualattentionbased} guide models to prioritize the crucial regions of images or the key words of questions. These approaches are not concerned with compositional generalization issues.
Owing to the exceptional performance of the Transformer structure~\cite{vaswani2017transformer}, VLMs~\cite{tan2019lxmert,wang2023image,ma2023boostingtransformerpretrain} have made great progress in generic VQA tasks.
Furthermore, due to the advancements attained by extensive model scaling and pre-training techniques, the recently proposed multimodal large language models (MLLMs)~\cite{zhang2024mmllm} have demonstrated remarkable capabilities in several multimodal tasks.
Nonetheless, the challenges of compositional generalization still persist and require focused solutions~\cite{yang2024exploringCompositional, dahlgren2024learningreasoning}.
In addition, few approaches focus on specialized VQA tasks with different settings.
For example, knowledge-based methods \cite{yang2022empirical,xu2024learningkbvqa} permit their models to access external knowledge during inference.
A recent study~\cite{agarwal2020towardscausalvqa} has put forward new metrics to measure the inconsistent predictions produced by a VQA model due to erroneous priors.
These specifically tailored methods can be applied to special VQA tasks and do not consider compositional VQA.

The prevalent VQA models focus primarily on the model construction process and training volume while overlooking the challenges of compositional generalization. In contrast, our emphasis is on a novel learning framework that purposively enhances the reasoning ability of the constructed model in the context of compositional VQA.

\subsection{Compositional VQA}
Some studies have attempted to address the generalization problems in VQA tasks, resulting in notable performance improvements.
Multiple methods leverage further parsed results obtained from inputs to adaptively adjust the computational processes of their models.
An earlier approach, FiLM~\cite{perez2018film}, introduces a feature-wise affine transformation to selectively manipulate the features of the network.
LR-Capsule~\cite{cao2021lrcapsule}, NS-VQA~\cite{yi2018nsvqa,dahlgren2023compositionalgeneralization}, and MGN~\cite{saqur2020mgn} parse the inputs and then conduct reasoning according to the parsed structural information.
Inspired by NMNs \cite{andreas2016nmn}, modular methods
\cite{bahdanau2019closure,d2021howmodular,chen2021mmn,yamada2022tmn, liu2024detection} decompose a question into subtasks and address each subtask with a specific module under the guidance of functional programs.
TbD \cite{mascharka2018tbd} is a modular network based on visual attention that closes the gap between performant and interpretable models.
These methods have made impactful contributions to explicit interpretable reasoning.
However, these models are conventionally trained without compositional variations, and they strongly rely on additional guidance information.
Recently, leveraging pre-training techniques, the model of MDETR~\cite{kamath2021mdetr} is pre-trained on image-text pairs to fulfill alignment requirements, and \cite{li2023exploring} directly leverages pre-trained VLMs to modify the original samples.
In addition, from the data distribution perspective, LexSym~\cite{akyurek2023lexsym} utilizes lexical symmetries for data augmentation purposes.
The improvements provided by these approaches primarily stem from pre-training or data processing rather than purposively addressing the root issues of understanding composable concepts and compositional shifts.

In conclusion, the current methods are not designed according to the essence of compositional variations, which lies in the changes occurring in the underlying concepts of input compositions.
We focus on enabling concept representation in a disentangled manner and learning a reasoning process with varying compositions through a novel training framework.

\subsection{Disentangled Representation and Equivariance}
Disentangled representation learning (DRL) is a crucial strategy for enabling models to extract the various properties of objects.
A common assumption in existing DRL studies is that latent factors should be statistically independent~\cite{wang2024disentangledsurvey}. A study~\cite{suter2019robustly} states that disentanglement of latent factors can be achieved by capturing the underlying causality~\cite{pearl2009causal}, and introduces a unifying causal framework to illustrate the issue of disentanglement. 
Based on the insight of~\cite{suter2019robustly}, recent research~\cite{reddy2022oncausally} further proposed the properties that DRL models should satisfy.  
Inspired by causality, several DRL methods~\cite{2021causalvae, Pal_2019, Disentangled2021} have proposed their assumptions of the causal structures of latent variables and attempted to model the underlying causal relationships. However, existing methods typically require extra detailed annotations to accomplish disentanglement, and rarely integrate the disentangled representations with downstream tasks. On the contrary, we propose a self-supervised intervention-based method for disentanglement and validly incorporate it into a novel framework for compositional VQA.

In a general context, equivariance is characterized as a property whereby the transformation of the input triggers a consistent and predictable transformation of the output. As such, it is well-suited for the research investigating the input-output correlations of neural networks.
Group equivariant CNN~\cite{cohen2016equivariantcnn} extends the translation equivariance of CNNs to further symmetry classes, which is regarded as the earliest work introducing equivariance to deep neural networks. Since then, equivariance has been utilized in several related fields.
Equivariant GNNs~\cite{2020GemNet} have been developed to better characterize the geometry and topology of geometric graphs. 
Subsequent studies~\cite{Lenssen_Fey_Libuschewski_2018, qi2020learningequivariant} make efforts on seeking representations that are equivariate to various transformations on images.
Although equivariance is now widely applied, no prior work has proposed a unified approach to address compositional generalization issues through equivariance.
Recent work~\cite{li2022equivariant} compels the model to identify relevant video frames through the equivariance between video inputs and answers, which seems to be the only work close to our approach. However, owing to
the discrepancy in the research questions, it is infeasible to
directly edit the implicit concepts as well as control the
output. Therefore, we design a novel framework that incorporates a new transformation method along with a specialized constraint measure.

\begin{figure}[t]
	\centering
	\includegraphics[width=0.65\linewidth]{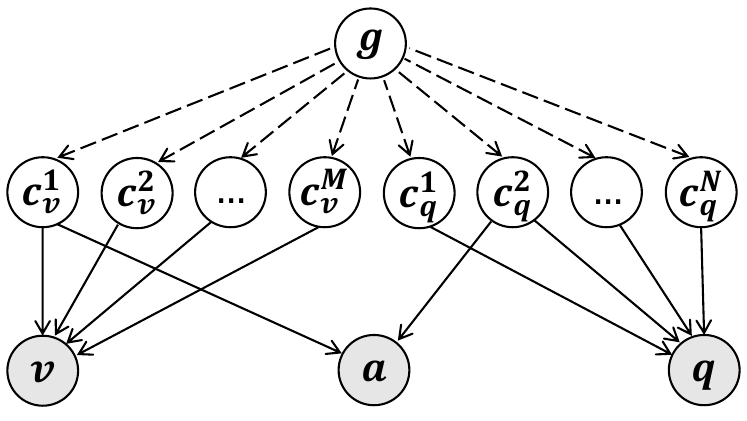}
	\caption{Causal graph of the variables involved in the VQA task. The concepts $\{\boldsymbol{c}_v^{i}\}_{i=1}^M$ and $\{\boldsymbol{c}_q^{i}\}_{i=1}^N$ are the causes of an image $\boldsymbol{v}$ and a question $\boldsymbol{q}$, respectively. Partial concepts are the causes of the answer $\boldsymbol{a}$. The confounder $\boldsymbol{g}$ arises from the underlying statistical biases in the datasets. The variables with gray shadows are observed, whereas the other variables are latent.}
    \label{fig_causalgragh}
\end{figure}

\section{Preliminaries}
\subsection{Problem Statement}
In the compositional VQA task, it is assumed that there exists a latent concept set $\mathcal{C}$ encompassing all visual and linguistic concepts in the dataset. 
In this dataset, each input image $\boldsymbol{v}$ and question $\boldsymbol{q}$ can be regarded as a visual concept composition and a linguistic concept composition, denoted as $\mathcal{C}_v$ and $\mathcal{C}_q$ respectively, where $\mathcal{C}_v, \mathcal{C}_q\subset \mathcal{C}$.
Given $\boldsymbol{v}$ and $\boldsymbol{q}$, a VQA model $f_\theta(\cdot)$ is required to predict the correct answer $\boldsymbol{\hat{a}}$ from the set of all candidate answers in the dataset~$\mathcal{A}$, \emph{i.e.},
$\boldsymbol{\hat{a}} = \arg\max_{\boldsymbol{a}\in \mathcal{A}}p_\theta (\boldsymbol{a}|\mathcal{C}_v,\mathcal{C}_q)$.

It is worth noting that the VQA model can access all concepts in $\mathcal{C}$ during training, but it is tested on novel unseen compositions of concepts from $\mathcal{C}$, which distinguishes it from the generic VQA task.
For instance, a model trained on input compositions $\left \langle c_1, c_3 \right \rangle$ and $\left \langle c_2, c_3 \right \rangle$ where $c_1,c_2,c_3 \in \mathcal{C}$, should be tested on a sample with composition $\left \langle c_1, c_2 \right \rangle$ that is not existing during training.

\subsection{A Causal View of Compositional VQA}\label{section_CausalViewonVQA}
A causal graph is constructed in Fig.~\ref{fig_causalgragh} to illustrate the causal mechanism of the VQA task, where a causal relationship is denoted as “\emph{Cause}$\longrightarrow $\emph{Effect}”.
Observed variables are contained in the graph, namely, an image $\boldsymbol{v}$, a question $\boldsymbol{q}$, and an answer $\boldsymbol{a}$. $M$ visual concepts and $N$ linguistic concepts are denoted as latent variables $\mathcal{C}_v=\{\boldsymbol{c}_v^{i}\}_{i=1}^M$ and $\mathcal{C}_q= \{\boldsymbol{c}_q^{i}\}_{i=1}^N$, respectively.

Following the definition presented in~\cite{reddy2022oncausally}, latent concepts can be regarded as independent generative factors of VQA samples, which implies that both $\boldsymbol{v}$ and $\boldsymbol{q}$ are formed by the compositions of concepts via $\mathcal{C}_v \longrightarrow \boldsymbol{v}$ and $\mathcal{C}_q \longrightarrow \boldsymbol{q}$.
However, only partial concepts could be the cause of $\boldsymbol{a}$, whereas other concepts are not causally related, \emph{e.g.}, $\boldsymbol{c}_v^{1}, \boldsymbol{c}_q^{2} \longrightarrow \boldsymbol{a}$ and $ \boldsymbol{c}_v^{2} \not\longrightarrow \boldsymbol{a}$.
As mentioned in~\cite{reddy2022oncausally}, considering the statistical biases existing in the generation process of VQA samples, there could be a confounder $\boldsymbol{g}$ which has noncausal effects on concepts, as shown in $\boldsymbol{g}\dashrightarrow \mathcal{C}_v,\mathcal{C}_q$. The current VQA approaches adopt the ubiquitous paradigm of empirical risk minimization (ERM) to optimize their models~\cite{tan2019lxmert,kim2021vilt}. Unfortunately, through the ERM paradigm, spurious statistical correlations caused by $\boldsymbol{g}$ might be captured~\cite{arjovsky2019invariant}.

As an illustration, the concepts “\emph{sphere}” ($\boldsymbol{c}_v^{1}$) and “\emph{yellow}” ($\boldsymbol{c}_v^{2}$) should be independent, but they could be confounded by $\boldsymbol{c}_v^{2}\dashleftarrow\boldsymbol{g} \dashrightarrow\boldsymbol{c}_v^{1}$ when $\langle$“\emph{sphere}”, “\emph{yellow}”$\rangle$ appears frequently throughout the training process.
When test questions query the color of a haphazardly colored sphere, $\boldsymbol{c}_v^{2}$ should not be related to the answer $\boldsymbol{a}$. However, a model trained under the ERM might be misled by $\boldsymbol{c}_v^{2}\dashleftarrow \boldsymbol{g} \dashrightarrow \boldsymbol{c}_v^{1}\longrightarrow \boldsymbol{a}$ and tend to predict “\emph{yellow}” with statistical prejudice. From the perspective of feature extraction, the semantics of different concepts are entangled in the latent space, which implies that $\boldsymbol{c}_v^{2}$ is inevitably involved in reasoning as long as $\boldsymbol{c}_v^{1}$ is demanded by the given question. Therefore, concept disentanglement is crucial for the reasoning process.
In our approach, we construct a novel re-encoding process to guarantee the independence of each concept by preventing it from being influenced by variations in other concepts.

\section{The Proposed Method}

\begin{figure*}[t]
    \centering
    \includegraphics[width=0.85\textwidth]{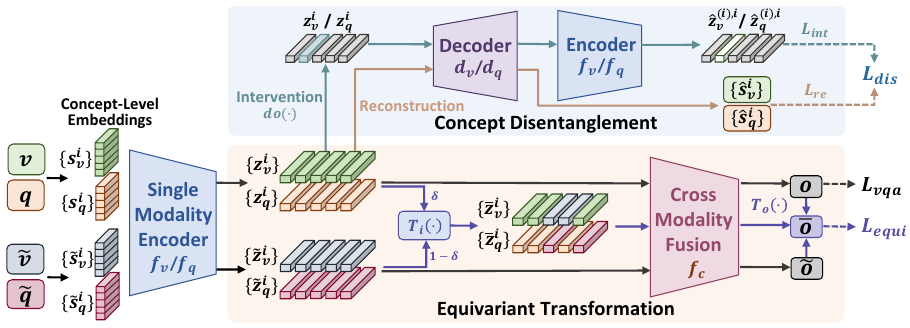}
    \caption{Overview of the proposed DEAL framework. In the proposed concept disentanglement module, causal interventions are applied to concepts, and each vector is forced to be unaffected by others; thus, each concept can be disentangled. Then, compositional transformations are performed in the latent space, and the corresponding outputs are constrained to satisfy equivariance. Thus, the model is encouraged to learn to reason novel compositions.}
    \label{fig_framework}
\end{figure*}

The overall architecture of our proposed DEAL framework is shown in Fig.~\ref{fig_framework}. Our unique designs primarily lie in two aspects: 
\textbf{concept disentanglement} and \textbf{equivariant transformation}.
Following the prevalent construction of VQA models, the networks implemented in the DEAL framework is built upon Transformer blocks~\cite{vaswani2017transformer}, and the pipeline involves two stages: a single-modality encoding process and a cross-modality reasoning process.
At the beginning, the inputs are converted into concept-level embeddings. 
Then, the concept disentanglement is performed during the single-modality encoding process to obtain disentangled representations, enabling the model to understand independent variation factors. Subsequently, the equivariant transformation is applied during the reasoning process to capture the mechanism of reasoning about compositional variations.

\subsection{Concept-level Embedding} 
Initially, $\boldsymbol{v}$ and $\boldsymbol{q}$ are converted into two comcept-level embedding sequences, denoted as $\mathcal{S}_v = \{\boldsymbol{s}_q^i\}_{i=1}^M$ and $\mathcal{S}_q = \{\boldsymbol{s}_v^i\}_{i=1}^N$ respectively, where $\boldsymbol{s}_q^1$ is the learnable global embedding. 
Specifically, for language inputs, each word token is treated as an individual concept in our framework.
For the visual branch, object extraction \cite{anderson2018bottom} is employed to obtain region-level embeddings, from which multiple concept-level embeddings can be further extracted for each region.

\subsection{Concept Disentanglement}\label{section_Concept_Factorization}
For the encoding process, the visual and language embeddings can be encoded as latent features by a visual encoder $f_v(\cdot)$ and language encoder $f_q(\cdot)$:
\begin{equation}\label{encoding_equation}
\mathcal{Z}_v = f_v(\mathcal{S}_v),\ \mathcal{Z}_q=f_q(\mathcal{S}_q),
\end{equation}
where $\mathcal{Z}_v=\{\boldsymbol{z}_v^i\}_{i=1}^M$ and $\mathcal{Z}_q=\{\boldsymbol{z}_q^i\}_{i=1}^{N}$. 
As discussed in Section~\ref{section_CausalViewonVQA}, with the ERM paradigm, concepts are always entangled in the latent dimensions due to noncausal paths via the confounder, \emph{i.e.}, the dashed edges in Fig.~\ref{fig_causalgragh}.
Hence, these paths should be purposively eliminated during the encoding process. As proposed in the theory of causality~\cite{pearl2009causal},
any extrinsic adjustment of a variable could be regarded as a causal intervention, which can cut off all incoming edges of the variable. For this reason, causal intervention is the prime measure used to disconnect spurious correlations.

\noindent{\bfseries Causal Intervention for VQA}: 
As mentioned above, taking Fig.~\ref{fig_causalgragh} as an illustration,
any intervention on $\boldsymbol{c}^j$ can remove its incoming edges, \emph{i.e.}, $\boldsymbol{g} \dashrightarrow \boldsymbol{c}^j$. As a consequence, the spurious noncausal connections are cut off, \emph{i.e.}, $\boldsymbol{c}^j \not\dashleftarrow \boldsymbol{g} \dashrightarrow \boldsymbol{c}^i \dashrightarrow \boldsymbol{v}$ (or $\boldsymbol{q}$, $\boldsymbol{a}$) ($i\neq j$).
Subsequently, other concepts, such as $\boldsymbol{c}^i$, no longer have effect on $\boldsymbol{v}$ (or $\boldsymbol{q}$, $\boldsymbol{a}$) via $\boldsymbol{c}^j$.
Following this property, we devise a unique operation to eliminate spurious correlations in our framework.
To obtain a disentangled representation of $\boldsymbol{c}^i_v$,
we keep $\boldsymbol{c}^i_v$ unchanged, whereas the other concepts are changed.
This leads to $\boldsymbol{g} \not\dashrightarrow \mathcal{C}_{in(i)}$, where $\mathcal{C}_{in(i)}$ = $\{\boldsymbol{c} | \boldsymbol{c}\in\mathcal{C}_v\cup\mathcal{C}_q,\boldsymbol{c}\ne \boldsymbol{c}^i_v\}$.
Consequently, from the angle of causality, the edge $\boldsymbol{c}^i_v\longrightarrow\boldsymbol{v}$ becomes the only path from $\boldsymbol{c}^i_v$ to $\boldsymbol{v}$, and it can be modeled independently.

For the fulfillment of the causal intervention property, we refer to~\cite{reddy2022oncausally}, which clarified that causal disentanglement can be interpreted by a generative autoencoder process and that each true generative factor $\boldsymbol{c}^i$ should be represented by unique latent dimensions $\boldsymbol{z}^i$. 
It is feasible to implement the proposed intervention operation, which can theoretically eliminate spurious correlations in such an autoencoder structure.
Conducting interventions on generative factors $\{\boldsymbol{z}^j\}$ ($i \neq j$) serving as proxies of $\{\boldsymbol{c}^j\}$ can be utilized as a measure for disconnecting the incoming edges of $\{\boldsymbol{c}^j\}$. Then, a disentangled $\boldsymbol{c}^i$, represented by $\boldsymbol{z}^i$, should be unaffected by changes in the other concepts and remain invariant throughout the encoding and decoding processes.

Build upon the above inspiration, an encoder (\emph{i.e.}, $f_v(\cdot)$ or $f_q(\cdot)$) and a corresponding decoder (\emph{i.e.}, $d_v(\cdot)$ or $d_q(\cdot)$) are employed in our disentanglement method, and the intervention operation is applied in the proposed novel re-encoding framework, as depicted in the “Concept Disentanglement” module of Fig.~\ref{fig_framework}.
Initially, the encoder and decoder must be trained by a prevailing reconstruction objective to learn the bijection between the latent space and the input embeddings (as shown in the “Reconstruction” branch).
After concept-level features are obtained by the encoding process, we execute the intervention by applying variations to the features other than a specified feature and then reserve the invariance of the specified feature during the subsequent decoding and encoding processes (as shown in the “Intervention” branch).
As long as a concept can remain unaffected by the changes in other concepts throughout the re-encoding process, disentanglement can be realized. The proposed method is detailed as follows.

\subsubsection{Reconstruction} 
For the question input, the $f_q(\cdot)$ and $d_q(\cdot)$ perform as autoencoder:
\begin{equation}
\mathcal{L}_{re\_ l}=\left\| d_q(\mathcal{Z}_q)- \mathcal{S}_q \right\|_{2}^{2}.
\end{equation}

Unlike the question input, visual concepts from images are continuous, and thus the reconstruction process for vision is designed following Variational Autoencoders (VAE)~\cite{kingma2013autovae}. 
The encoder and decoder are optimized by minimizing the following objective:
\begin{equation}
\mathcal{L}_{re\_ v}= -\mathbb{E}_{q(\boldsymbol{z}_v|\mathcal{S}_v)}\log p\left(\mathcal{S}_v|\boldsymbol{z}_{v}\right) +  KL(q(\boldsymbol{z}_v|\mathcal{S}_v)\|p(\boldsymbol{z}_v)),
\end{equation}
where the first term is equivalent to the reconstruction loss, and the second term is KL divergence between the posterior distribution $q(\boldsymbol{z}_v|\mathcal{S}_v)$ and prior distribution $p(\boldsymbol{z}_v)$. The total reconstruction loss is $\mathcal{L}_{re}= \mathcal{L}_{re\_ l}+\mathcal{L}_{re\_ v}$. However, with $\mathcal{L}_{re}$ as the sole objective, the underlying concepts are still entangled in the latent space.

\subsubsection{Intervention}
Based on the decoders and encoders that learn the generative mechanism, we proceed with the proposed intervention operation.
As previously introduced, the proposed operation $\operatorname{do}(\cdot)$ is implemented by retaining one concept feature while altering the others.
Specifically, for the question input, $\operatorname{do}(\cdot)$ applied on $\mathcal{Z}_q$ can be formulated as
\begin{equation}\label{eq:intervene language}
\mathcal{Z}_q^{(i)} = \operatorname{do}(\mathcal{Z}_q,i) 
=\{\boldsymbol{\epsilon}^1,\ldots, \boldsymbol{\epsilon}^{i-1},\boldsymbol{z}^i,\boldsymbol{\epsilon}^{i+1},\ldots,\boldsymbol{\epsilon}^N\},
\end{equation}
where $\boldsymbol{z}_q^i$ is the unchanged feature, and the vectors $\{\boldsymbol{\epsilon}^j | j\in[1,N],j\ne i\}$ are randomly sampled from encoded features of other samples.
Next, the intervened features are decoded as $d_q(\mathcal{Z}_q^{(i)})$, and then put into the encoder to obtain the re-encoded output: 
\begin{equation}\label{eq:reencoding_language}
\hat{\mathcal{Z}}_q^{(i)} =f_q(d_q(\mathcal{Z}_q^{(i)})). 
\end{equation}

Since $\boldsymbol{c}_q^i$ should be the only concept retained from the original input until the end, the $i\rm th$ feature vector of the re-encoded output, \emph{i.e.} $\boldsymbol{\hat{z}}_q^{(i),i}$, should approximate $\boldsymbol{z}_q^i$. This optimization is further performed on each feature of $\mathcal{Z}_q$. Formally:
\begin{equation}\label{eq:disentangle language}
\mathcal{L}_{int\_l}= \mathbb{E}_{\boldsymbol{z}_q^i \in \mathcal{Z}_q} \left\| \boldsymbol{\hat{z}}_q^{(i),i}-\boldsymbol{z}_q^i \right\|_{2}^{2}.
\end{equation}

For the visual input, similarly to Equations~\ref{eq:intervene language} and~\ref{eq:reencoding_language}, the intervened visual features $\mathcal{Z}_{v}^{(i)}=\operatorname{do}(\mathcal{Z}_{v},i)$ and the re-encoded output $\hat{\mathcal{Z}}_{v}^{(i)}=f_v(d_v(\mathcal{Z}_v^{(i)}))$ can be obtained.
The optimization objective is to make  the $i\rm th$ feature of $\hat{\mathcal{Z}}_{v}^{(i)}$ approximate corresponding $\boldsymbol{z}_v^i$:
\begin{equation}\label{eq:disentangle vision} 
\mathcal{L}_{int\_v}= \mathbb{E}_{\boldsymbol{z}_v^i \in \mathcal{Z}_v} \left\| \boldsymbol{\hat{z}}_v^{(i),i}-\boldsymbol{z}_v^i \right\|_{2}^{2}.
\end{equation}

The total loss of intervention becomes $\mathcal{L}_{int}=\mathcal{L}_{int\_v}+\mathcal{L}_{int\_l}$. Finally, the optimization objective of concept disentanglement can be written as $\mathcal{L}_{dis} =\mathcal{L}_{re}+ \mathcal{L}_{int}$.

\subsection{Equivariant Transformation}
\label{subsection Equivariant Transformation for VQA}

After the encoding process of $\mathcal{Z}_v$ and $\mathcal{Z}_q$, the model proceeds with cross-modality reasoning with these latent features.
The multimodal encoder $f_c(\cdot)$ is employed to output the fused global feature $\boldsymbol{o}_q^1$ which is then attached to a classifier to predict the answer $\boldsymbol{\hat{a}}$: 
\begin{equation}\label{reasoning_equation}
\begin{aligned}
    \boldsymbol{o}_q^1 = f_c(\mathcal{Z}_v, \mathcal{Z}_q), \ \boldsymbol{\hat{a}}={\rm Classifier}(\boldsymbol{o}_q^1).
\end{aligned}
\end{equation}

A novel image-question input can be perceived as a transformation from another input sample by changing some disentangled concepts.
Equivariance is often used to investigate the alterations in the output produced by a neural network according to the input variations~\cite{cohen2016equivariantcnn, qi2020learningequivariant, li2022equivariant}. Thus, we propose an equivariance-based framework to train the model with input concept variations.

\noindent{\bfseries Equivariance for Compositional VQA}:
As depicted in the “Equivariant Transformation” module in Fig.~\ref{fig_framework}, $\mathcal{T}_i(\cdot)$ and $\mathcal{T}_o(\cdot)$ are defined as the transformations of the input and output spaces of $f_c(\cdot)$, which should comply with the principle of equivariance.
As long as $\mathcal{T}_i(\cdot)$ applied to the input of $f_c(\cdot)$, the equivariant variation $\mathcal{T}_o(\cdot)$ must arise in the output of $f_c(\cdot)$:
\begin{equation}\label{equation_equivariance}
f_c(\mathcal{T}_i(\mathcal{Z}_c)) =\mathcal{T}_o(f_c(\mathcal{Z}_c)),
\end{equation}
where the cross-modality representation $\mathcal{Z}_c = \{\boldsymbol{z}_c^i\}_{i=1}^{M+N}$ is used to represent $\mathcal{Z}_v \cup \mathcal{Z}_q$ for the sake of simplicity. 
Concretely, as disentangled features
are proxies of underlying concepts, $\mathcal{T}_i(\cdot)$ is distinctively devised as imposing changes on the latent features to create unseen compositions.
Then, the model can be trained by reasoning the correct answers with novel input compositions of varied seen concepts and the equivariant constraint. The details regarding our method are presented below.

\subsubsection{Compositional Transformation}
In our method, to prepare for $\mathcal{T}_i(\cdot)$, another randomly sampled input pair, $\boldsymbol{\tilde{v}}$ and $\boldsymbol{\tilde{q}}$, is processed through the VQA model as given in Equations~\ref{encoding_equation} and~\ref{reasoning_equation}; thus, $\tilde{\mathcal{Z}}_c$ and $\boldsymbol{\tilde{o}}_q^1$ can be obtained.

We define $N_{t}$ as the number of features in $\mathcal{Z}_c$ which are randomly selected to be replaced by features from $\tilde{\mathcal{Z}}_c$. For instance, if $N_{t}=2$, the new composition $\bar{\mathcal{Z}}_c$ can be generated by:
\begin{equation}\label{Ti example}
\bar{\mathcal{Z}}_c = \mathcal{T}_i(\mathcal{Z}_c)
=\{\boldsymbol{z}_c^1,\boldsymbol{\tilde{z}}_c^2,\boldsymbol{\tilde{z}}_c^3,\boldsymbol{z}_c^4,\ldots,\boldsymbol{z}_c^{M+N}\},
\end{equation}
where the features $\boldsymbol{z}_c^2, \boldsymbol{z}_c^3$ are randomly chosen to be replaced by $\boldsymbol{\tilde{z}}_c^2, \boldsymbol{\tilde{z}}_c^3$. We define $\mathcal{Z}_{trans}$ as the set of replaced features (\emph{e.g.}, in Equation~\ref{Ti example}, $\mathcal{Z}_{trans}=\{\boldsymbol{z}_c^2, \boldsymbol{z}_c^3\}$).
Afterward, the reasoning output becomes
\begin{equation}\label{equa_Ti_output}
\boldsymbol{\bar{o}}_q^1=f_c(\mathcal{T}_i(\bar{\mathcal{Z}_c})).
\end{equation}

\subsubsection{Equivariant Constraint}
According to Equation~\ref{equation_equivariance}, when $\mathcal{T}_i(\cdot)$ is applied, the output of $f_c(\cdot)$ should satisfy equivariant transformation $\mathcal{T}_o(\cdot)$. A recent study~\cite{li2022equivariant} proposed a training framework for video question answering utilizing linear interpolation as the equivariant constraint. Specifically, they set a hyperparameter as the transformation ratio, and
the ratio of changed input frames equals the transformation ratio of the output.
Similarly in our case, the constraint is applied to the output global embedding ${\boldsymbol{o}}_q^1$:
\begin{equation}\label{Eq:transformation of output}
\mathcal{T}_o(f_c(\mathcal{Z}_c)) = \mathcal{T}_o({\boldsymbol{o}}_q^1) =  \delta\boldsymbol{ {o}}_q^1 + (1-\delta) \boldsymbol{\tilde{o}}_q^1 ,
\end{equation}
where $\delta$ represents the transformation ratio that implies the degree of transformation from $\mathcal{Z}_c$ to $\bar{\mathcal{Z}}_c$. 
However, $\delta$ cannot be decided simply like counting the number of input frames.
Therefore, a unique measure is demanded to properly quantify the transformation ratio $\delta$ instead of setting hyperparameters.

Aiming to quantify $\delta$, we first define $\{\lambda_i\}_{i=1}^{M+N}$ as the importance weights of concept features of $\mathcal{Z}_c$, since transformations of different features have varying effects. Following Grad-CAM~\cite{selvaraju2017grad}, the weight of $\boldsymbol{z}_c^i$ is calculated as:
\begin{equation}\label{Eq:Grad-CAM}
\lambda_{i}=\frac{1}{N_d}  \sum_{j=1}^{N_d} \frac{\partial p_\theta (\boldsymbol{a}|\boldsymbol{v},\boldsymbol{q})}{\partial z_c^{i,j}},
\end{equation}
where $N_d$ denotes the dimensionality of each vector and $z_c^{i,j}$ is the $j{\rm th}$ variable of $\boldsymbol{z}_c^{i}$; $p_\theta (\boldsymbol{a}|\boldsymbol{v},\boldsymbol{q})$ represents the model's prediction score for the ground-truth answer $\boldsymbol{a}$.
Similarly, $\tilde{\lambda}_i$ can be calculated for each $\boldsymbol{\tilde{z}}_c^i$. Then the balanced weights are obtained by $\bar{\lambda}_i=(\lambda_i+\tilde{\lambda}_i)/2$.

Subsequently, by measuring the distance between each pair of features in $\mathcal{Z}_c$ and $\tilde{\mathcal{Z}}_c$, $\delta$ can be determined by calculating the proportion of the total distance caused by the replacement:
\begin{equation}\label{transformscale}
    \delta=\frac{\sum_{\boldsymbol{z}_c^i \in \mathcal{Z}_{trans}}\bar{\lambda}_i \left\|\boldsymbol{z}_c^i- \boldsymbol{\tilde{z}}_c^i \right\|_{2}^{2}}{{\sum_{j=1}^{M+N}}\bar{\lambda}_j \left\|\boldsymbol{z}_c^j- \boldsymbol{\tilde{z}}_c^j\right\|_{2}^{2}}.
\end{equation}

Finally, based on Equations~\ref{equa_Ti_output} and~\ref{Eq:transformation of output}, the objective of equivariant constraint can be formulated as:
\begin{equation}\label{Eq:Equivariant Constraint}
\mathcal{L} _{equi} = \left \| \mathcal{T}_o(f_c(\mathcal{Z}_c)) -f_c(\mathcal{T}_i(\mathcal{Z}_c)) \right \|_2^2.
\end{equation}

Experimental results are provided in Sections~\ref{sec_ablation} and~\ref{sec_validationofmodules} to validate the effectiveness and design intentions of our proposed disentanglement and equivariant learning methods.

\subsection{Optimization}
Following the typical VQA methods, here we apply binary cross-entropy ${\rm BCELoss}(\cdot)$ on the prediction $\boldsymbol{\hat{a}}$:
\begin{equation}
    \mathcal{L}_{vqa}= {\rm BCELoss}( \boldsymbol{a},\boldsymbol{\hat{a}}).
\end{equation}
To sum up, the overall training objective of DEAL is the aggregation of all aforementioned objectives:
\begin{equation}
\label{total_loss}
\mathcal{L}_{DEAL}=\mathbb{E}_{(\boldsymbol{v},\boldsymbol{q},\boldsymbol{a}) \in \mathcal{D}_{train}} (\mathcal{L}_{vqa}+\alpha \mathcal{L}_{dis}+\beta \mathcal{L} _{equi}),
\end{equation}
where $\mathcal{D}_{train}$ is the training set; $\alpha$ and $\beta$ are hyperparameters that balance the strength of each training objective.

\section{Experiments}

\subsection{Datasets}
We validate the compositional generalization ability of our proposed DEAL framework on two public datasets: \textbf{CLEVR-CoGenT}~\cite{johnson2017clevr} and \textbf{GQA-SGL}~\cite{yamada2022tmn}.
Additionally, two conventional VQA datasets, \textbf{CLEVR}~\cite{johnson2017clevr} and \textbf{VQAv2.0}~\cite{goyal2017making}, are utilized to showcase the model's performance in generic scenarios.

CLEVR-CoGenT is a variant of CLEVR~\cite{johnson2017clevr}, and it is a diagnostic dataset specifically designed for testing the compositional generalization capability of a VQA model. It contains $130,000$ synthetic 3D images, with approximately $10$ automatically generated questions per image. These images contain various simple objects with randomly generated attributes such as shapes, colors, and materials. There are two parts: Condition A and Condition B. In Condition A, the cubes are gray, blue, brown, or yellow, and the cylinders are red, green, purple, or cyan. In Condition B, the situation is interchanged. By experimenting on CLEVR-CoGenT, we can evaluate the visual compositional generalization ability of a VQA model.

The GQA balanced split is a dataset derived from GQA~\cite{hudson2019gqa} that features a more balanced answer distribution to mitigate language priors.
It is designed to evaluate reasoning skills using 22,000 natural images and 140,000 questions. The GQA balanced split includes a training set and a test-dev set for evaluation purposes.
GQA-SGL~\cite{yamada2022tmn} is a test set created based on the test-dev set of the GQA balanced split, and it contains images from the GQA balanced split along with $200$ novel questions belonging to four types (verify, query, choose, and logical). These questions are generated with ground-truth programs by combining existing attributes and referring expressions; thus, the GQA-SGL test samples possess novel linguistic compositions that do not exist in GQA. The ability of a model to perform linguistic compositional generalization can be evaluated on GQA-SGL.

CLEVR is the original version of CLEVR-CoGenT, which is commonly utilized to evaluate complex visual reasoning abilities. It contains approximately one million automatically generated questions and rendered images with 3D objects. VQAv2.0 is one of the most widely used VQA benchmarks. It includes training, validation, and test splits, comprising a total of approximately $1.1$ million image-question pairs with approximately $13$ million associated answers for approximately $200$k images from COCO~\cite{Zitnick_2014}. The data for training and testing in each of these two datasets share the same distribution.

\noindent{\bfseries Evaluation Metric}: All the experiments are conducted in the form of open-ended VQA, which signifies that the output space of the model comprises all the possible answers in the dataset. A prediction is considered correct when the model output matches the ground-truth answer, and the evaluation metric is the prediction accuracy of the model across all test samples.

\subsection{Implementation Details}

Same as prevailing visual-language models, the input question is projected to 
$N$ word-level $768$-dimensional embeddings by pre-trained BERT \cite{devlin2018bert}, where $N$ is $47$ for CLEVR-CoGenT and $20$ for GQA-SGL.
For visual input,  the pre-trained  Faster R-CNN \cite{ren2015faster} is applied to extract $12$ objects along with position features following \cite{anderson2018bottom}, and then MLPs are used to extract $5$ concept-level $768$-dimensional embeddings for each object. 
Next, embeddings of vision and language are sent into the proposed model $f_\theta(\cdot)$ for prediction.
The backbone of $f_\theta(\cdot)$ is constructed based on Transformer \cite{vaswani2017transformer} with the uniform hidden dimensionality $N_d = 768$. 
The structure and numbers of the layers in single modality encoders ($f_q(\cdot)$ and $f_v(\cdot)$) and the cross modality encoder ($f_c(\cdot)$) are build up following LXMERT \cite{tan2019lxmert} which is a representative Transformer-based model.
Subsequently, a two-layer MLP with hidden size $1,536$ is utilized as the answer head. 
For concept disentanglement, the decoders $d_q$ and $d_v$ are structured the same as $f_q$ and $f_v$. 
For the equivariant transformation, a positive integer $\eta=4$ is defined as the number of iterations we apply compositional transformations during training on one sample. 
In the first iteration, $N_t = \lfloor (M+N) /\eta\rfloor$ features are randomly selected to be altered, and in each subsequent iteration, $N_t$ additional features are added to be transformed each time.

Our experiments are conducted with the computational resource of two A100 GPUs.
BertAdam optimizer~\cite{devlin2018bert} is used for all cases. The hyperparameters $\alpha$ and $\beta$ are set to $0.03$ and $0.1$ for both CLEVR-CoGenT and GQA-SGL. Learning rates for CLEVR-CoGenT and GQA-SGL are set to $1e-5$ and $2e-5$ respectively, The experimental setups conducted on the CLEVR and VQAv2.0 are identical to the experiments on CLEVR-CoGenT and GQA-SGL, respectively.

\subsection{Comparison with the SOTA Methods}\label{comparison_exp}
To validate the superiority of our proposed DEAL approach in the compositional VQA task, we provide experimental results and quantitative analyses from four perspectives.
1)
Experiments are conducted on CLEVR-CoGenT to compare DEAL with the SOTA methods when generalizing to novel visual compositions.
2) Experiments are conducted on GQA-SGL to compare DEAL with the SOTA methods when generalizing to novel linguistic compositions.
3) To intuitively display the generalization ability of the evaluated models, the difference between its in-distribution and generalization performances is exhibited and analyzed.
4) To verify that DEAL maintains competitive performance in a generic in-distribution environment, we present the experimental results obtained on CLEVR and VQAv2.0, even though DEAL is not intended for contributing to in-distribution performance.

The experimental results are listed in Table~\ref{tab:CLEVR-CoGenT}, Table~\ref{tab:GQA-SGL}, and Table~\ref{tab:CLEVRVQAv2}, where the “In-dist.” and “Generalization” columns represent the results obtained under the in-distribution and compositional generalization settings, respectively.
The values with “$\downarrow$” symbols in parentheses indicate the gap between the in-distribution and generalization performances.
The “Add.” column implies that the corresponding method benefits from additional assistance, including pre-training (PT), functional programs (PR), and linguistic parse trees (PA).

\noindent{\bfseries Compared Methods}: 
 Several leading compositional VQA approaches are cited for comparison, including 6 approaches that adaptively adjust model computations  (FiLM, NS-VQA, TbD-reg, MMN, TMN-Tree, and LR-Capsule) and 4 Transformer-based models (MDETR, LXMERT, ViLT, and LexSym). 
\textbf{FiLM}~\cite{perez2018film} enables a neural network to perform adaptive reasoning by applying feature-wise affine transformations conditioned on input.
The model of \textbf{NS-VQA}~\cite{yi2018nsvqa} is developed under programs and scene representations as additional training signals, and requires the programs as guidance for inference.
Approaches based on modular networks, \textbf{TbD-reg}~\cite{mascharka2018tbd}, \textbf{MMN}~\cite{chen2021mmn}, and \textbf{TMN-Tree}~\cite{yamada2022tmn}, initially decompose questions by programs and then tackle sub-tasks with different modules. Both the training and testing processes rely on ground-truth programs.
\textbf{LR-Capsule} uses the question parse as extra tools tree to generate adaptive reasoning routines within the capsule network. 
\textbf{MDETR} \cite{kamath2021mdetr} 
leverages $1.3$ million image-text pairs to capture the alignment between image regions and words during pre-training. 
Mainstream VLMs, \textbf{LXMERT}~\cite{tan2019lxmert} and \textbf{ViLT}~\cite{kim2021vilt}, pre-train their multimodal networks on multiple datasets including GQA balanced split, and are commonly applied on the generic VQA task.
\textbf{LexSym}~\cite{akyurek2023lexsym} is an augmentation-based approach based on data symmetry without requirements for pre-training or additional assistance.

\noindent{\bfseries Visual Compositional VQA}:
As there is a visual compositional gap between Condition A and Condition B, we conduct experiments on CLEVR-CoGenT to demonstrate that DEAL exhibits optimal performance in visual compositional generalization scenarios.
All the models are trained solely on the training set of Condition A. Afterward, their in-distribution performance is evaluated on the validation set of Condition A, and their generalization performance is evaluated on the validation set of Condition B.
The experimental results obtained on CLEVR-CoGenT are listed in Table~\ref{tab:CLEVR-CoGenT}.
Compared with FiLM and LexSym, which also do not necessitate any additional assistance, DEAL possesses a significant advantage in both Conditions A and B, particularly outperforming them by $9.4$\% and $2.3$\%, respectively, under the generalization setting.
Modular models, including TbD-reg, NS-VQA, LR-Capsule, and TMN-Tree, are trained with additional supervision to guide their reasoning processes. We surpass the best-performing LR-Capsule method by $2.6$\%, demonstrating that our approach holds a reasoning advantage in cases with compositional variations.
Transformer-based models such as MDETR, ViLT, and LXMERT exhibit strong in-distribution capabilities, benefiting from abundant pre-training knowledge. MDETR even implements a specific alignment-based pre-training process for cross-modal understanding.
However, most VQA datasets do not provide
additional auxiliary information, and
the performance of the competitors patently decreases when generalizing to Condition B.
Consequently, the DEAL model maintains decent in-distribution performance and meanwhile manifestly outperforms the other approaches when generalizing to novel visual compositions.

\begin{table}[t]
\small
\centering
\caption{Answering accuracy achieved on the CLEVR-CoGenT validation sets of Condition A (in-distribution) and Condition B (compositional generalization). All models used for testing are solely trained on Condition A.}
\label{tab:CLEVR-CoGenT}  
  \begin{tabular}{l|c|c|c}\hline
    Methods&\makecell[c]{Add.}&\makecell[c]{CoGenT A\\ (In-dist.)}& \makecell[c]{$\textbf{CoGenT B}$\\(Generalization)}\\\hline
    FiLM (0-Shot) \cite{perez2018film} & - &98.3&78.8 (19.5$\downarrow$)\\
    NS-VQA \cite{yi2018nsvqa}  & PR &99.8&63.9 (35.9$\downarrow$)\\
    TbD+reg \cite{mascharka2018tbd} & PR &98.8&75.4 (23.4$\downarrow$)\\
    LR-Capsule \cite{cao2021lrcapsule} & PA &98.1&85.6 (12.5$\downarrow$)\\
    TMN-Tree \cite{yamada2022tmn}& PR &98.0$\pm$0.02 & 80.1$\pm$0.72 (17.9$\downarrow$)\\\hline
    MDETR \cite{kamath2021mdetr}& PT & $\textbf{99.8}$& 76.7 (23.1$\downarrow$)\\
    ViLT \cite{kim2021vilt}& PT &98.2& 83.3 (14.9$\downarrow$)\\
    LXMERT \cite{tan2019lxmert}& PT &99.1& 86.7 (12.4$\downarrow$)\\
    LexSym \cite{akyurek2023lexsym}& - &-& 85.9$\pm$0.9(-)\\\hline
    $\textbf{DEAL}$ &-  &99.2 $\pm$ 0.4 & $\textbf{88.2$\pm$0.4 (10.9$\downarrow$)}$ \\\hline
\end{tabular}
\end{table}

\begin{table}[t]
\small
\centering
\caption{Answering accuracy achieved on GQA test-dev (in-distribution) and GQA-SGL (compositional generalization ). All models used for testing are solely trained on the GQA balanced split.}
\label{tab:GQA-SGL}
  \begin{tabular}{l|c|c|c}\hline
    Methods&\makecell[c]{Add.}&\makecell[c]{GQA test-dev\\ (In-dist.)}&\makecell[c]{$\textbf{GQA-SGL test}$\\(Generalization)}\\\hline
    MMN \cite{chen2021mmn}&PR&67.8& 51.5$\pm$1.5 (16.3$\downarrow$)\\
    TMN-Tree \cite{yamada2022tmn}& PR &65.2 & 53.7$\pm$1.7 (11.5$\downarrow$)\\\hline
    LXMERT \cite{tan2019lxmert}& PT &68.9& 58.0 (10.9$\downarrow$)\\
    MDETR \cite{kamath2021mdetr}& PT & $\textbf{73.9}$& 59.0 (14.9$\downarrow$)\\
    ViLT \cite{kim2021vilt}& PT &65.8& 55.2 (10.6$\downarrow$)\\\hline
    $\textbf{DEAL}$ &-  &69.0 $\pm$ 0.7 & $\textbf{61.4$\pm$0.9 (6.8$\downarrow$)}$\\\hline
\end{tabular}
\end{table}

\noindent{\bfseries Linguistic Compositional VQA}:
The training set of the GQA balanced split has the same compositions as the GQA test-dev set does, but it possesses linguistic compositions that are entirely distinct from those of GQA-SGL.
Therefore, the models are tested on GQA-SGL to validate the superiority of DEAL in terms of addressing linguistic compositional generalization.
Moreover, the results obtained on GQA-SGL also verify the VQA reasoning capabilities of DEAL for real images.
All the models are trained only on the GQA balanced split and then tested on the GQA test-dev set and GQA-SGL to evaluate their in-distribution and generalization performances, respectively.
The results of the experiments conducted on GQA-SGL are listed in Table~\ref{tab:GQA-SGL}.
Modular models such as MMN~\cite{chen2021mmn} and TMN utilize ground-truth programs that can only parse linguistic inputs; hence, they fail to achieve ideal performance on datasets with more complex natural images. Compared with them, our approach shows advantages on both test sets, particularly outperforming TMN by $7.7$\% under the generalization condition.
With the advantage of extensive pre-training data, MDETR, LXMERT, and ViLT achieve better results compared to the other modular methods.
Among them, MDETR has an advantage over the other approaches, especially in terms of in-distribution performance, as it not only utilizes pre-trained encoding modules but also specifically conducts an additional pre-training process for cross-modal alignment. Notably, the images used for its pre-training process and the samples in GQA test-dev are sourced from the same domain, \emph{i.e.}, COCO~\cite{Zitnick_2014}. Nonetheless, the model exhibits a performance decrease when faced with linguistic generalization. Without extensive pre-training, our DEAL method still
maintains its in-distribution performance. In addition, in the context of linguistic compositional settings, our model achieves SOTA performance.

\begin{table}[t]
\small
\centering
\caption{Answering accuracy achieved on CLEVR and VQAv2.0. The training and testing samples share the same data distribution without compositional generalization.}
\label{tab:CLEVRVQAv2}  
  \begin{tabular}{l|c|c|c}\hline
    Methods&Add.&\makecell[c]{CLEVR\\ (In-dist.)}&\makecell[c]{VQAv2.0\\ (In-dist.)}\\\hline
    LexSym \cite{akyurek2023lexsym}& - &92.0 $\pm$ 0.9& 67.8 $\pm$ 0.7 \\
    FiLM \cite{perez2018film} &-&97.7& - \\
    LR-Capsule \cite{cao2021lrcapsule} & PA &-&67.0\\
    NS-VQA \cite{yi2018nsvqa} &PR& $\textbf{99.8}$& 62.2 $\pm$ 0.3 \\\hline
    ViLT \cite{kim2021vilt} &PT& 98.2 $\pm$ 0.5 &71.3\\
    LXMERT \cite{tan2019lxmert}  &PT& 99.2 $\pm$ 0.2  &72.5\\
    MDETR \cite{kamath2021mdetr} &PT& 99.7& $\textbf{73.8 $\pm$ 0.3}$ \\\hline
    $\textbf{DEAL}$ &-  &99.4 $\pm$ 0.2 & 73.1 $\pm$ 0.4\\\hline
\end{tabular}
\end{table}

\noindent{\bfseries Decline in Generalization}:
The gap between the performances attained by the same model under the in-distribution and generalization settings provides an intuitive indication of the generalization ability of the model.
An optimal model should present the slightest decline when generalizing to a new distribution.
The results in Tables~\ref{tab:CLEVR-CoGenT} and~\ref{tab:GQA-SGL} show that the existing approaches, including MDETR, LXMERT, and NS-VQA, have advanced in-distribution performance. Their design intentions concentrate on improving the training volume or interpretability by making full use of additional training assistance. Nonetheless, they also present evident performance decreases when encountering generalization, \emph{e.g.}, MDETR on CLEVR-CoGenT ($23.1$\%) and GQA test-dev ($14.9$\%), LXMERT on CLEVR-CoGenT ($12.4$\%) and GQA test-dev ($10.9$\%), and NS-VQA on CLEVR-CoGenT ($35.9$\%).
Comparatively, in terms of both visual and linguistic generalization, our DEAL method yields the narrowest gaps, \emph{i.e.}, $10.9$\% on CLEVR-CoGenT and $6.8$\% on GQA test-dev. This finding indicates that our method of disentanglement and compositional equivariant transformation significantly improves the model's ability to understand novel compositions of seen concepts.

\noindent{\bfseries In-Distribution VQA}:
In CLEVR and VQAv2.0, the training and test splits share the same visual and linguistic compositions.
Hence, experiments are conducted on CLEVR and VQAv2.0 to validate whether DEAL performs strongly under a generic in-distribution VQA setting, even though the contribution of DEAL does not concern in-distribution performance.
The experimental results are listed in Table~\ref{tab:CLEVRVQAv2}.
The results indicate that our approach outperforms most of the competitors and is only slightly inferior to MDETR and NS-VQA on CLEVR.
However, as previously noted, they both benefit from additional auxiliary information. MDETR conducts extensive pre-training for image-text alignment. In particular, the images used for pre-training are derived from the same source domain (COCO~\cite{Zitnick_2014}) as those of VQAv2.0. NS-VQA requires question programs and visual scene representations during training and even relies on programs during the inference process. Notably, most VQA data lack additional training information, and the reliance of these models on extra support limits their deployment in broad scenarios. In addition, their generalization performances are obviously inferior.
In conclusion, DEAL achieves SOTA performance with respect to compositional generalization of VQA while possessing competitive in-distribution performance that relies solely on ground-truth answers.

\begin{table*}[t]
\small
\centering
\caption{The results of ablation study conducted on CLEVR-CoGenT and GQA-SGL.}
\label{tab:Ablation}
\begin{tabular}{lccc|cc|cc}\hline
    \multirow{2}{*}{Methods}&\multirow{2}{*}{\thead{$\mathcal{L}_{vqa}$}}&\multirow{2}{*}{\thead{$\mathcal{L}_{dis}$}}&\multirow{2}{*}{\thead{$\mathcal{L}_{equi}$}}&\multicolumn{2}{c|}{CLEVR-CoGenT}&\multicolumn{2}{c}{GQA-SGL}\\
    &&&&  CoGenT A& CoGenT B&GQA test-dev& GQA-SGL test\\\hline
    ${\rm DEAL}_{\rm{w/o\;cd\&ec}}$& \ding{51}&&&98.2 $\pm$ 0.2&84.3 $\pm$ 0.3&67.7 $\pm$ 0.3&54.6 $\pm$ 0.9\\
    ${\rm DEAL}_{\rm w/o\;ec}$& \ding{51}& \ding{51} & &98.3 $\pm$ 0.3& 85.2 $\pm$ 0.9 &67.8 $\pm$ 0.7& 56.0 $\pm$ 0.8\\
    ${\rm DEAL}_{\rm w/o\;cd}$& \ding{51}&& \ding{51}&98.8 $\pm$ 0.5& 87.1 $\pm$ 0.4&68.1 $\pm$ 0.4& 59.1 $\pm$ 0.4\\
    $\textbf{DEAL (Ours)}$& \ding{51}& \ding{51} & \ding{51} &\textbf{99.2 $\pm$ 0.4}& $\textbf{88.2 $\pm$ 0.4}$&\textbf{69.0 $\pm$ 0.7}& $\textbf{61.4 $\pm$ 0.9}$  \\\hline
\end{tabular}

\end{table*}

\subsection{Ablation Study}\label{sec_ablation}
To analyze the effectiveness of the proposed training objectives of DEAL (\emph{i.e.}, $\mathcal{L}_{dis}$ and $\mathcal{L}_{equi}$) under in-distribution and compositional generalization settings, we further conduct an ablation study involving DEAL on CLEVR-CoGenT and GQA-SGL.
For CLEVR-CoGenT, the model is trained on Condition A and tested individually on Conditions A and B. For GQA-SGL, the model is trained on the GQA balanced split and tested individually on the GQA test-dev and GQA-SGL test set.
The results are listed in Table~\ref{tab:Ablation}.
The test accuracy of the baseline model ${\rm DEAL}_{\rm{w/o\;cd\&ec}}$ trained with only a ubiquitous VQA loss is shown in the first row.
Next, the results produced by the model trained without equivariant constraints ${\rm DEAL}_{\rm{w/o\;ec}}$ or concept disentanglement ${\rm DEAL}_{\rm{w/o\;cd}}$ are listed in the following rows.
Across the two test conditions of the two datasets (corresponding to the four columns of Table~\ref{tab:Ablation}), the experimental outcomes generally validate that both $\mathcal{L}_{dis}$ and $\mathcal{L}_{equi}$ are beneficial to the performance of the model.
Specifically, the results obtained under the in-distribution setting show that $\mathcal{L}_{dis}$ and $\mathcal{L}_{equi}$ yield performance improvements of $0.1$\% and $0.6$\%, respectively, on CLEVR-CoGenT, and improvements of $0.1$\% and $0.4$\%, respectively, on the GQA-SGL test set. Under the generalization setting, $\mathcal{L}_{dis}$ and $\mathcal{L}_{equi}$ yield improvements of $0.9$\% and $2.8$\%, respectively, on CLEVR-CoGenT, and improvements of $1.4$\% and $4.5$\%, respectively, on the GQA-SGL test set.
The improvements become more pronounced when $\mathcal{L}_{dis}$ and $\mathcal{L}_{equi}$ are applied simultaneously, as illustrated in the last row.
Therefore, both the disentangled concept representation and equivariant learning methods contribute to enhancing the reasoning capabilities of the model, whether under the in-distribution or compositional generalization condition.

\subsection{Validation of the Proposed Modules}\label{sec_validationofmodules}

As introduced in the motivation for DEAL, our first objective is to develop a concept disentanglement module that can make concept features relatively invariant, which implies that a concept feature should not be influenced by changes in other features. Second, the learning approach of equivariant transformation is expected to enable the model to infer corresponding outputs for unseen transformed compositions.
In this section, experiments are conducted to evaluate whether these design goals are achieved.

\noindent{\bfseries Concept Disentanglement}:
In our proposed method, concept disentanglement is achieved through the constraint $\mathcal{L}_{int}$, which is based on causal intervention.
To verify the effectiveness of $\mathcal{L}_{int}$, two models are needed for comparison: one trained without $\mathcal{L}_{int}$ and
another trained with $\mathcal{L}_{int}$. During testing, for both models, we assess whether a specified concept feature is affected by the alteration of other features. Ideally, the visual and linguistic concept features encoded by the model trained with $\mathcal{L}_{int}$ possess better relative invariance, indicating that the concepts are successfully disentangled.

Concretely, to evaluate whether the visual and language features are disentangled, the trained models are tested on $1000$ images and $1000$ questions randomly sampled from CLEVR-CoGenT.
Then, we train first the model with $\mathcal{L}_{vqa}$ and $\mathcal{L}_{re}$, and then train the second model with an additional objective $\mathcal{L}_{int}$. For the first model, each question input can be converted into features $\mathcal{Z}_q=\{\boldsymbol{z}_q^i\}_{i=1}^N$ by $f_q(\cdot)$. Following the re-encoding process formulated as Equations~\ref{eq:intervene language} and~\ref{eq:reencoding_language}, we randomly select one feature for intervention $\operatorname{do}(\mathcal{Z}_q,i)$ and obtain the re-encoded features $\hat{\mathcal{Z}}_q^{(i)}$. Then, we quantify the changes in the $i\rm th$ feature before and after the entire re-encoding process: $\omega = \left\| \boldsymbol{\hat{z}}_q^{(i),i}-\boldsymbol{z}_q^i \right\|_{2}^{2}$. Similarly to the aforementioned calculation process, we calculate $\omega$ for each image input. 
At this point, each $\omega$ of each input image and question sample is obtained for the first model.
After then, following the same calculation process of $\omega$, each $\tilde{\omega}$ can be obtained for each image and question sample for the second model.

\begin{figure}[t]
	\centering
	\includegraphics[width=0.83\linewidth]{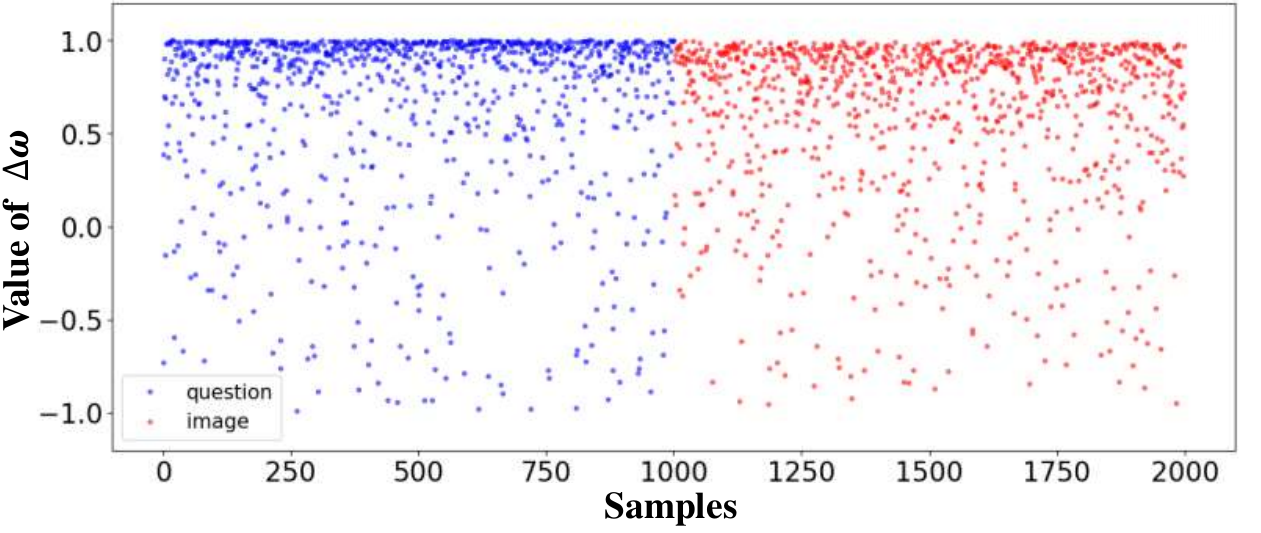}
	\caption{Analysis of the concept disentanglement module. Each tick on the horizontal axis represents a sample selected for validation, and each point represents the calculated value of $\Delta\omega$ for the corresponding sample. $\Delta\omega>0$ implies that the examined concept is better disentangled with the help of $\mathcal{L}_{int}$ in the concept disentanglement module.}
    \label{fig_validation_cd}
\end{figure}

It's worth noting that a lower value of $\omega$ (or $\tilde{\omega}$) implies that the $i\rm th$ feature is less affected by the changes in other features, \emph{i.e.}, the $i\rm th$ feature is better disentangled from others. To compare the values of $\omega$ and $\tilde{\omega}$, we plot $\Delta\omega=\tanh(\omega-\tilde{\omega})$ for each image and question sample, as shown in Fig.~\ref{fig_validation_cd}. The results illustrate that $\Delta\omega> 0$ for most samples, suggesting that the second model trained with $\mathcal{L}_{int}$ generally maintains a relatively lower value of $\tilde{\omega}$. This implies that the proposed module contributes effectively to concept disentanglement.

\begin{figure}[t]
	\centering
	\includegraphics[width=0.83\linewidth]{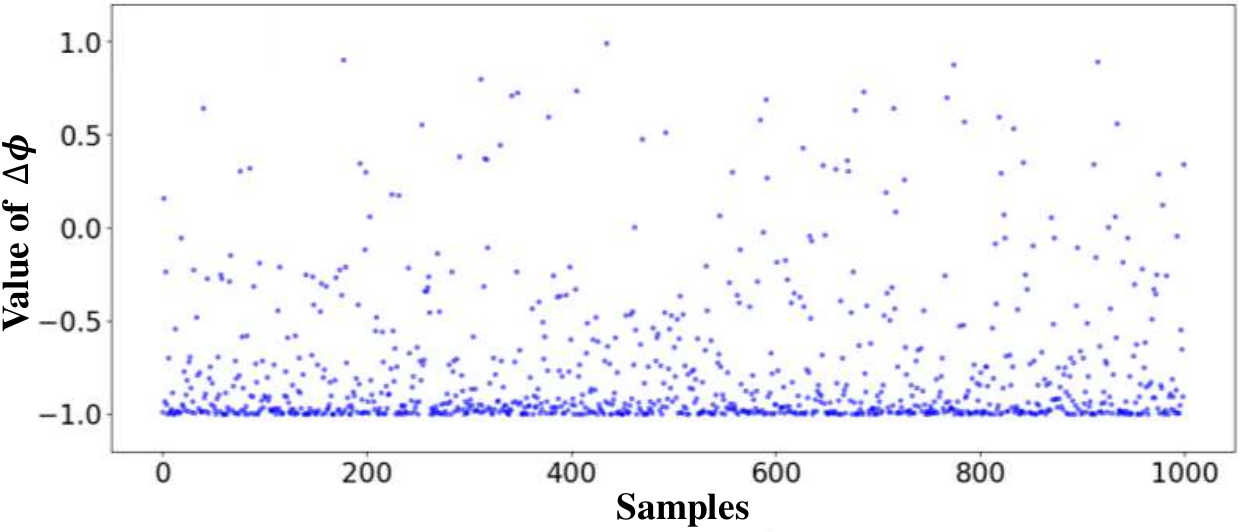}
	\caption{Analysis of the equivariant transformation module. Each tick on the horizontal axis represents a sample selected for validation, and each point represents the calculated value of $\Delta\phi$ for the corresponding sample. $\Delta\phi<0$ implies that the equivariant transformation implemented with $\mathcal{L}_{equi}$ facilitates the model in achieving equivariance.}
    \label{fig_validation_ec}
\end{figure} 

\noindent{\bfseries Equivariant Transformation}:
In our proposed equivariant learning approach, the model acquires its compositional reasoning ability by
incorporating concept feature alterations and equivariantly constraining the transformation of the corresponding output by $\mathcal{L}_{equi}$.
To validate the improvement brought by the learning approach, two models trained with different approaches are compared: one trained without the equivariant constraint $\mathcal{L}_{equi}$ and another trained with $\mathcal{L}_{equi}$.
These two models are tested separately to investigate whether they can output the corresponding results with unseen compositions as their inference inputs.

Specifically, we test the baseline model ${\rm DEAL}_{\rm{w/o\;cd\&ec}}$ and the model ${\rm DEAL}_{\rm{w/o\;cd}}$ trained by the additional $\mathcal{L}_{equi}$ on $1000$ randomly sampled image-question inputs, and the transformation $\mathcal{T}_i(\cdot)$ are uniformly applied for each input during inferring. For ${\rm DEAL}_{\rm{w/o\;cd}}$, given an input sample, $\delta$ can be obtained following Equation~\ref{transformscale}. Then we calculate $\phi = \left \|  \bar{\boldsymbol{o}} - (\delta\boldsymbol{o} + (1-\delta) \boldsymbol{\tilde{o}}) \right \|_2^2$ based on Equation~\ref{Eq:transformation of output}. 
So far each $\phi$ can be calculated for the corresponding input image-question pair for ${\rm DEAL}_{\rm{w/o\;cd}}$.
Next, for the baseline model ${\rm DEAL}_{\rm{w/o\;cd\&ec}}$, we calculate each $\tilde{\phi}$ in the same way as $\phi$.

Notably, a lower value of $\phi$ (or $\tilde{\phi}$) indicates that the model better satisfies equivariant transformation, \emph{i.e.}, the inference output changes properly according to the input concept variations. Thus we plot the value of each $\Delta\phi=\tanh(\phi-\tilde{\phi})$ for the corresponding image-question sample, as shown in Fig.~\ref{fig_validation_ec}. 
It can be easily observed that $\Delta\phi< 0$ for most samples, implying that
the model ${\rm DEAL}_{\rm{w/o\;cd}}$ trained with equivariant transformation and the constraint $\mathcal{L}_{equi}$ can better handle novel compositions. This verifies that the proposed equivariant learning approach effectively aids the model in compositional reasoning.

\begin{figure}[t]
\centering
\subfloat[]{%
\includegraphics[width=0.95\linewidth]{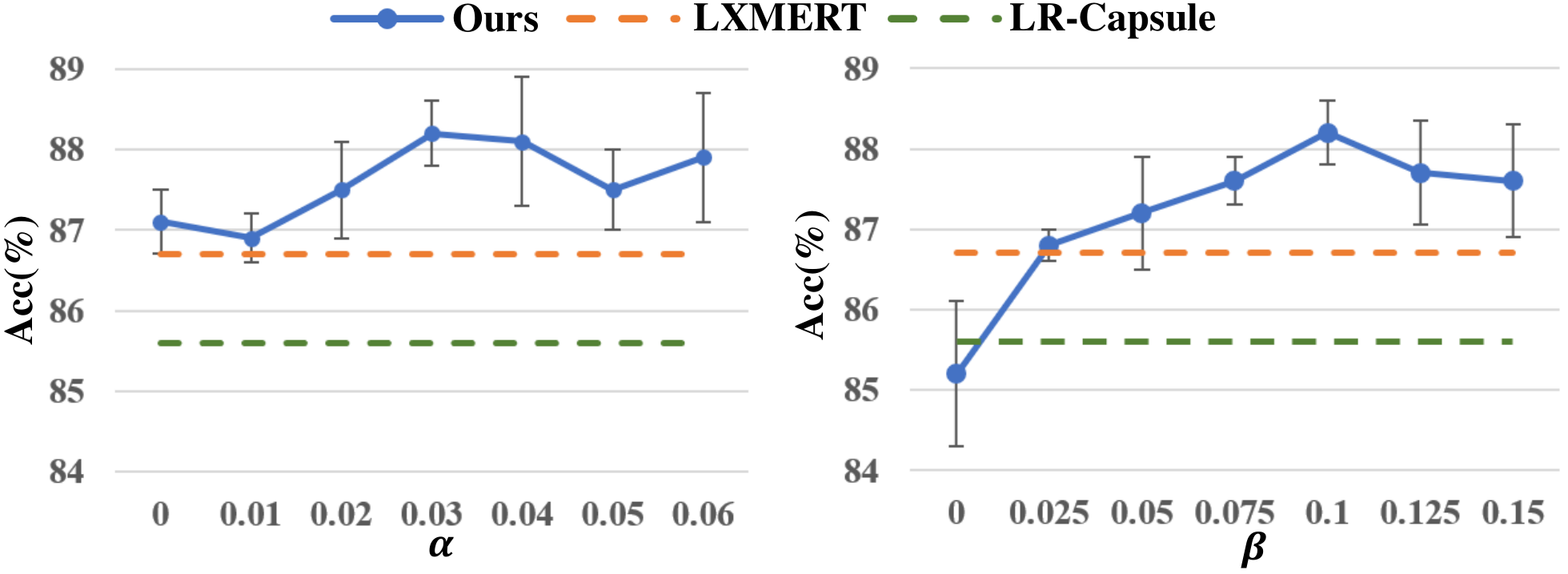}
}%
\quad %
\subfloat[]{%
\includegraphics[width=0.95\linewidth]{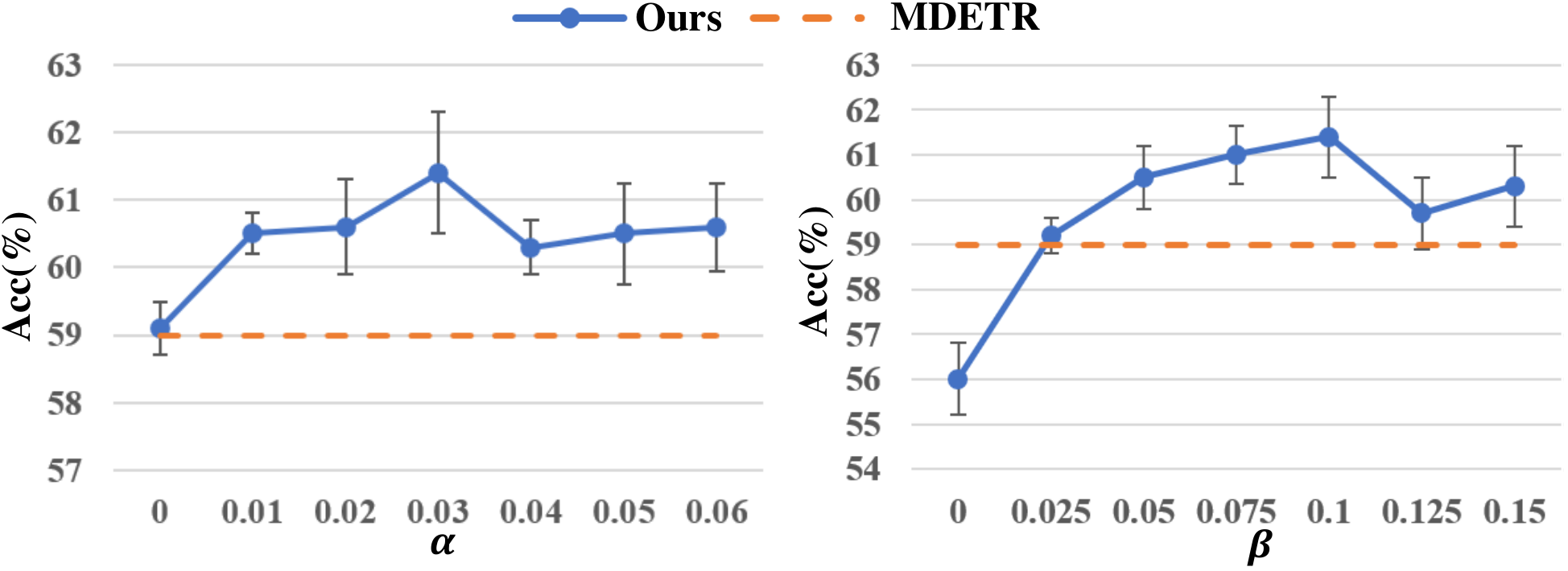}
}
\caption{Answering accuracy achieved on CLEVR-CoGenT Condition B (a) and the GQA-SGL test (b). Left: different $\alpha$ values with $\beta$ set to 0.1. Right: different $\beta$ values with $\alpha$ set to 0.03.}
\label{fig_hyper}
\end{figure}

\begin{figure*}[t]
\centering
\subfloat[]{
\includegraphics[width=0.405\linewidth]{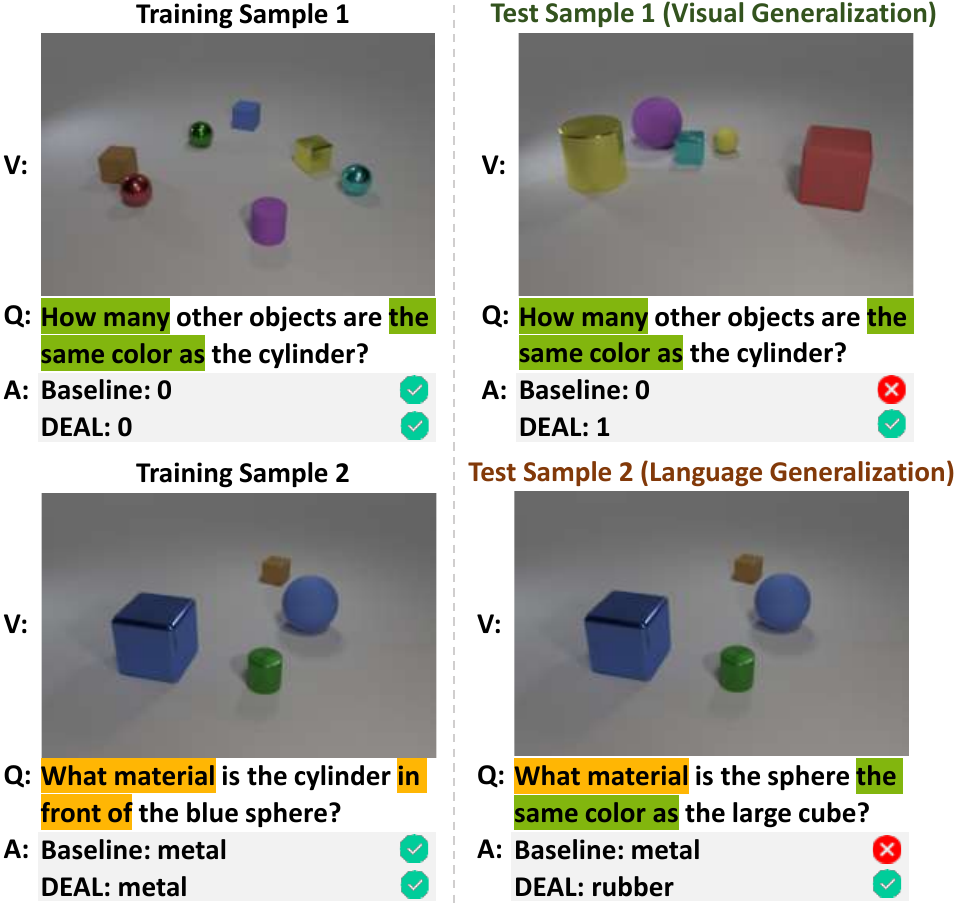}
}
\subfloat[]{
\includegraphics[width=0.405\linewidth]{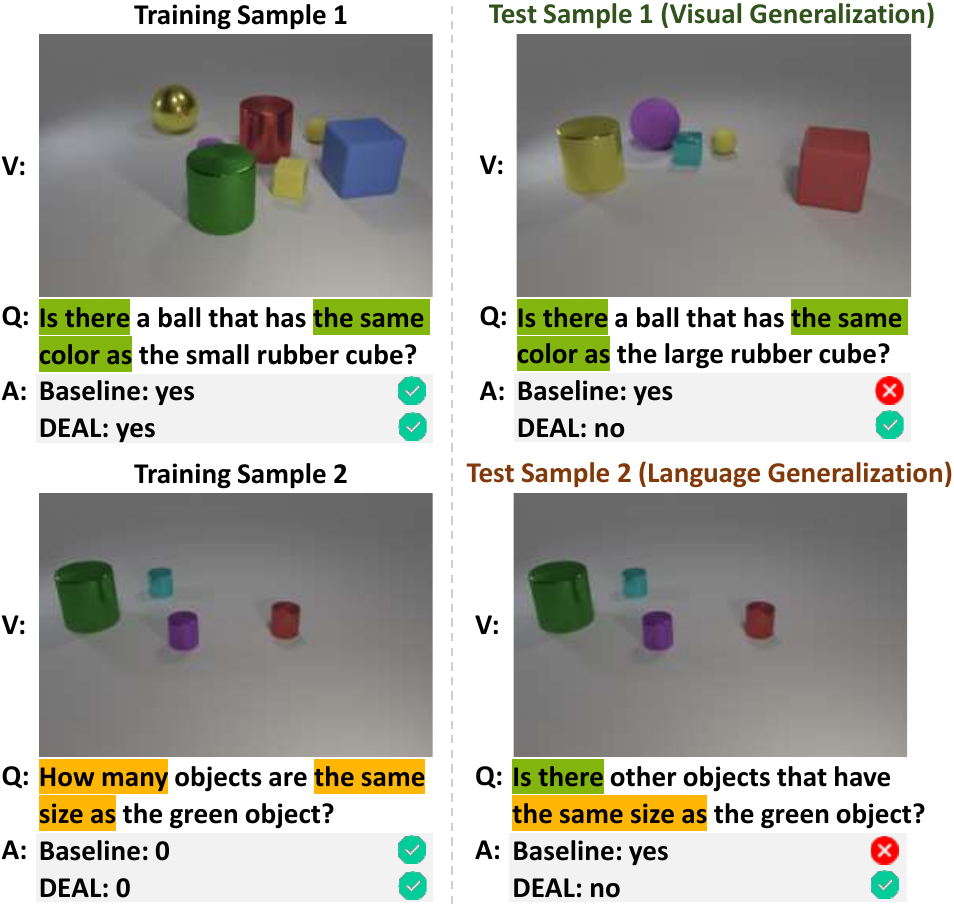}
}
\caption{Successful cases predicted by our approach. (a) and (b) are two groups of cases, where \emph{Test Sample 1} and \emph{Test Sample 2} in each group are used to test the compositional generalization capability of the model for vision and language, respectively.}
\label{fig_case}
\end{figure*} 

\begin{figure*}[!h]
\centering
\subfloat[]{
\includegraphics[width=0.18\linewidth]{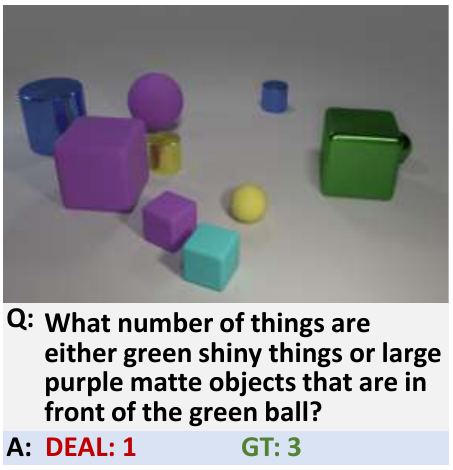}
}
\subfloat[]{
\includegraphics[width=0.18\linewidth]{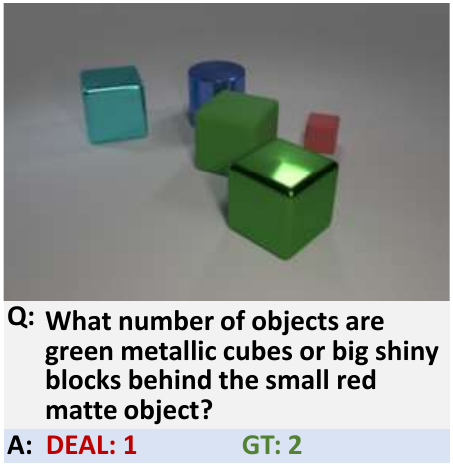}
}
\subfloat[]{
\includegraphics[width=0.18\linewidth]{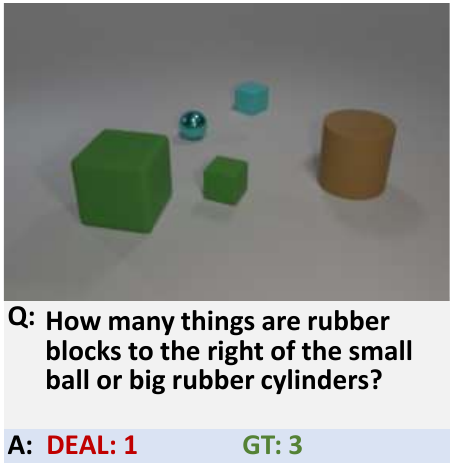}
}\\
\subfloat[]{
\includegraphics[width=0.18\linewidth]{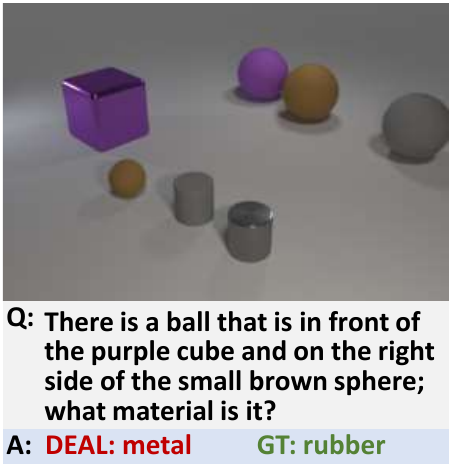}
}
\subfloat[]{
\includegraphics[width=0.18\linewidth]{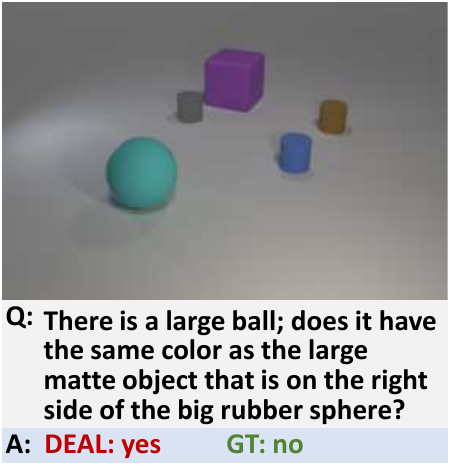}
}
\subfloat[]{
\includegraphics[width=0.18\linewidth]{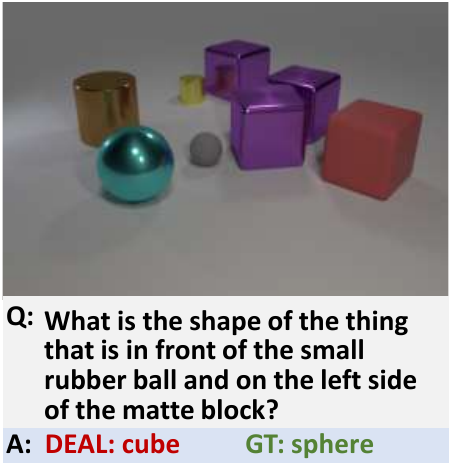}
}

\caption{Failure cases predicted by our approach. “GT” represents the ground-truth answer. (a), (b) and (c) are samples with the querying type of counting. (d), (e) and (f) are samples requiring multi-step reasoning.}
\label{fig_case_fail}
\end{figure*}

\subsection{Hyperparameter Analysis}
To analyze the impacts of the hyperparameters $\alpha$ and $\beta$ in Equation~\ref{total_loss} on the performance of DEAL, we conduct a hyperparameter analysis experiment.
Additionally, to verify whether DEAL can maintain its superiority across various hyperparameter values, several developed models are selected for comparison.
The model is trained with different values of $\alpha$ and $\beta$ and then evaluated under the compositional generalization setting, which is the same as that of the comparative experiments in Section~\ref{comparison_exp}.
$\alpha$ is adjusted within the range of $[0, 0.06]$ with steps of $0.01$, and $\beta$ is adjusted within $[0,0.15]$ with steps of $0.025$. For the test conducted on CLEVR-CoGenT, LR-Capsule and LXMERT are selected as the representative explicit reasoning and Transformer-based methods. For the test conducted on GQA-SGL, MDTER is taken as the reference, which benefits from pre-training to achieve strong performance.

The experimental results are shown in Fig.~\ref{fig_hyper}.
As can be observed from the results, increasing the weight of $\mathcal{L}_{dis}$ from zero has a beneficial effect on the achieved accuracy, and the performance is optimal when $\alpha=0.03$. This promotes feature disentanglement and thus facilitates compositional reasoning. $\mathcal{L}_{equi}$ has a more noticeable impact on the performance improvement, as it enables the model to encounter novel compositions during training. When $\beta=0.1$, the model achieves SOTA performance.
Moreover, except when the hyperparameters are set to 0, our model consistently outperforms the other methods as the hyperparameters change within their ranges.
These results indicate the strong advantages of our proposed DEAL approach and the robustness of the model to changes in its hyperparameters.

\subsection{Case Study}
\noindent{\bfseries Successful Cases and Analysis}: 
To offer more qualitative evidence of the advantage provided by DEAL for compositional VQA, we present some actual successful cases and further analyze the results on the basis of the design motivation of the proposed method.
In this case study, two models trained on the Condition A training set are evaluated, including the DEAL model and the baseline model trained with only the $\mathcal{L}_{vqa}$ objective.
As shown in Fig.~\ref{fig_case}, for both groups of examples, the VQA triplets in the left column are sampled from the Condition A training set, and the test triplets containing novel compositions in the right column are manually constructed by combining known concepts.
Specifically, \emph{Test Sample 1} comprises a question with the same linguistic constructs as those of \emph{Training Sample 1} and an image sampled from Condition B that exhibits novel visual compositions.
\emph{Test Sample 2} contains an image that is the same as \emph{Training Sample 2} and a question that presents novel linguistic compositions with seen components from \emph{Training Samples 1} and \emph{2}.
\emph{Test Sample 1} and \emph{Test Sample 2} are used to demonstrate the generalization ability of the model for vision and language, respectively.

In \emph{Test Sample 1} of Fig.~\ref{fig_case}(a), as cubes and cylinders have opposite colors under Conditions A and B, the model does not encountered “\emph{yellow cylinder}” during training.
Thus, the baseline model fails to understand that the “\emph{cylinder}” is “\emph{yellow}” in the image.
In contrast, with the help of the disentanglement module, our proposed DEAL approach is able to understand individual visual concepts, such as “\emph{cylinder}” and “\emph{yellow}”.
In \emph{Test Sample 2} of Fig.~\ref{fig_case}(a), the question is newly constructed based on “\emph{query for material}” and “\emph{color referring}”. Due to the entanglement of the baseline model, there might be spurious correlations between the “\emph{query for material}” and the answer “\emph{metal}”, and the model fails to comprehend the novel composition of the “\emph{query for material}” and the “\emph{color referring}”, leading to the incorrect output “\emph{metal}”. However, DEAL is trained to acquire an understanding of disentangled linguistic concepts to produce correct predictions.
Similarly, in Fig.~\ref{fig_case}(b), the baseline model misunderstands “\emph{red}” and “\emph{cube}”, as there is no “\emph{red cube}” composition during the training process conducted under Condition A; thus, it cannot correctly recognize the color of the “\emph{cube}” in \emph{Test Sample 1}. In comparison, our DEAL method can disentangle the concepts of colors and shapes and understand novel visual compositions such as “\emph{red}” and “\emph{cube}”.
In \emph{Test Sample 2}, with the input linguistic concept changes from “\emph{color referring}” to “\emph{size referring}”, the baseline model fails to adjust its answer correspondingly, leading to the same answer “\emph{yes}” as that produced for Training Sample 1, which is incorrect.
In contrast, owing to the equivariant learning approach, the DEAL model can correctly reason about novel linguistic compositions.

Consequently, this case study illustrates the superiority of our proposed concept disentanglement and equivariant learning methods in the compositional generalization task of VQA.

\noindent{\bfseries Failure Cases and Analysis}:
The DEAL framework aids the model in compositional VQA scenarios. However, some issues remain during testing. To visually highlight the shortcomings of DEAL, we present several typical failure cases from Condition B, as shown in Fig.~\ref{fig_case_fail}. In these cases, the DEAL model is tested on CLEVR-CoGENT with the compositional generalization setup. With the illustration of the results, we elaborate on their causes in this section.

Similar to the existing methods, the proposed model struggles with counting questions, as shown in Figs.~\ref{fig_case_fail}(a), (b), and (c). Owing to the larger quantities of counting and various referring types (\emph{e.g.}, spatial, color, and size) that might be incorporated into the questions, visual comprehension becomes more challenging, and reasoning becomes tricky. In addition, the reasoning process of our model is constructed holistically based on Transformer, which implies that DEAL cannot adaptively adjust its reasoning process for questions with varying lengths, types, and reasoning conditions. Consequently, DEAL has relatively inferior performance in complex reasoning tasks, especially when the questions require multiple reasoning steps, as shown in the cases presented in Figs.~\ref{fig_case_fail}(d), (e), and (f).

Methods employing adaptive modular reasoning hold an advantage in complex reasoning tasks. Thus, in future research, it would be feasible to integrate the merits of our framework and modular thinking, which could potentially address complex reasoning tasks more effectively.
Ideally, modular reasoning procedures would be conducted for inference purposes, and the model could utilize disentangled multimodal representations and be trained via the proposed learning framework.

\section{Conclusion}
In this paper, we focus on enhancing the compositional generalization ability of a VQA model, which presents a critical challenge for robust machine reasoning.
We introduce the viewpoint that the essence of compositional generalization in VQA lies in the compositional variations of independent generative factors, \emph{i.e.}, visual and linguistic concepts. Thus, the key problems are twofold: 1) how to extract disentangled factors from input image-question pairs and 2) how to enhance compositional reasoning capabilities by following the essence of compositional variations.
However, the existing methods fail to adequately address these challenges.
To this end, we propose a novel disentanglement-based equivariant learning framework for compositional VQA. In DEAL, causal interventions are applied in a re-encoding framework to enable the model's disentangled representation of concepts. Subsequently, feature-wise transformations are performed in the latent space, and the equivariant constraint is imposed on the output. Extensive experiments conducted on compositional VQA datasets validate that concept disentanglement and equivariant transformation both improve the compositional reasoning results and that DEAL outperforms other approaches when generalizing to novel compositions.
With respect to future research, DEAL has the potential for integration with other multimodal models. Specifically, modular networks can leverage our methods in DEAL as training procedures to enhance their reasoning capabilities, as previously discussed.
Moreover, our implementation and experimental results validate the effectiveness and superiority of our approach based on Transformer-based architectures.
Therefore, by integrating the advantages of the proposed concept disentanglement and equivariant learning methods, we are able to enhance the training and fine-tuning strategies of a wide range of Transformer-based models, such as various VLMs and MLLMs.
This improvement is expected to boost the models' capabilities in disentangled understanding of concepts and reasoning when confronted with compositional variations, which is a highly promising solution for handling compositional generalization in future research.

\bibliographystyle{IEEEtran}
\bibliography{cite}


\begin{IEEEbiography}
[{\includegraphics[width=1in, height=1.25in, clip, keepaspectratio]{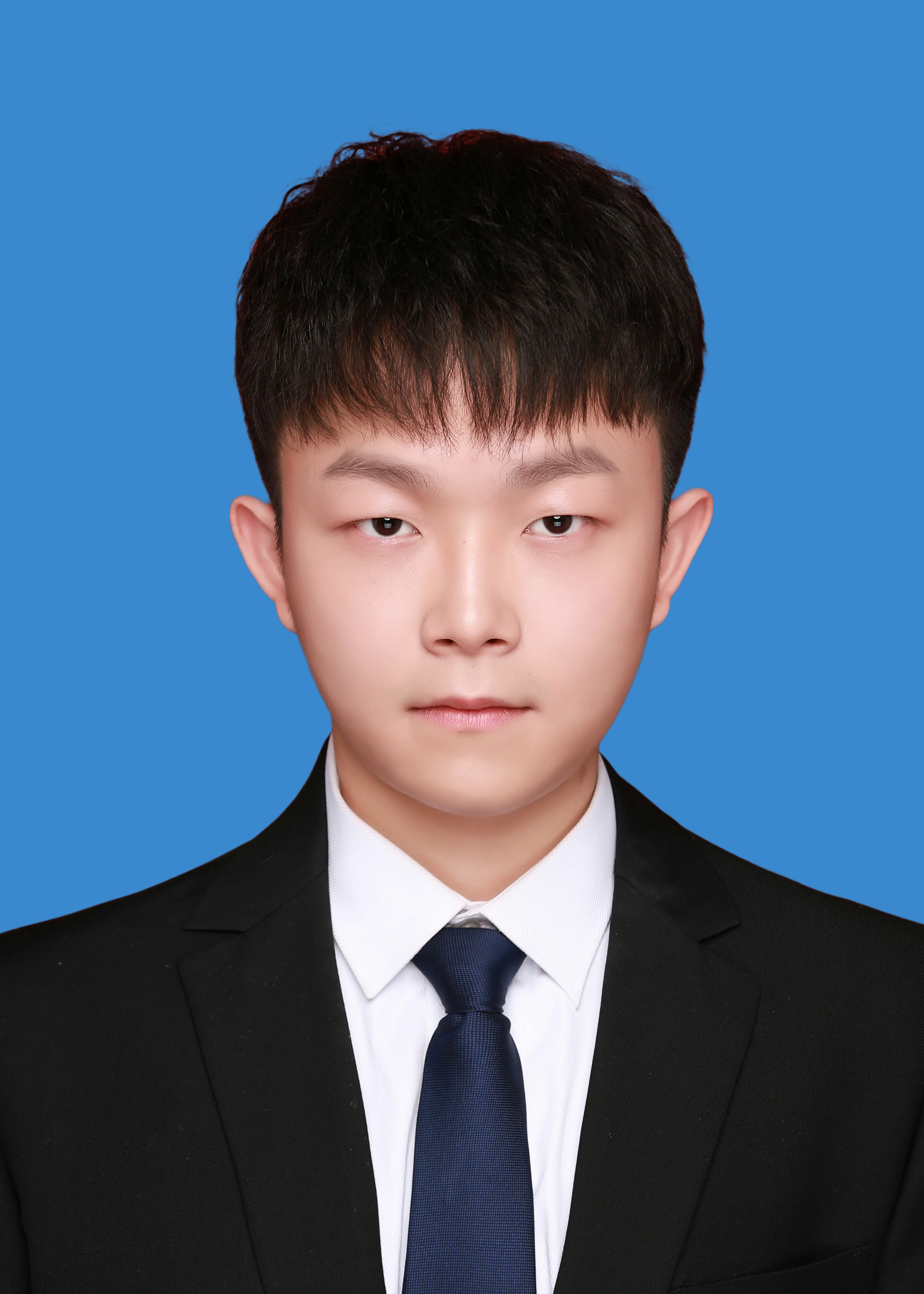}}]
{Zhou Du} received the B.E. degree at the School of Information Science and Technology from the Southwest Jiaotong University (SWJTU), Chengdu, China. He is currently working towards the M.S. degree at the School of Computing and Artificial Intelligence at Southwest Jiaotong University (SWJTU), Chengdu, China. His research interests include machine learning, computer vision, and multi-modal learning.
\end{IEEEbiography}

\vspace{11pt}
\vspace{-30pt}

\begin{IEEEbiography}
[{\includegraphics[width=1in, height=1.25in, clip, keepaspectratio]{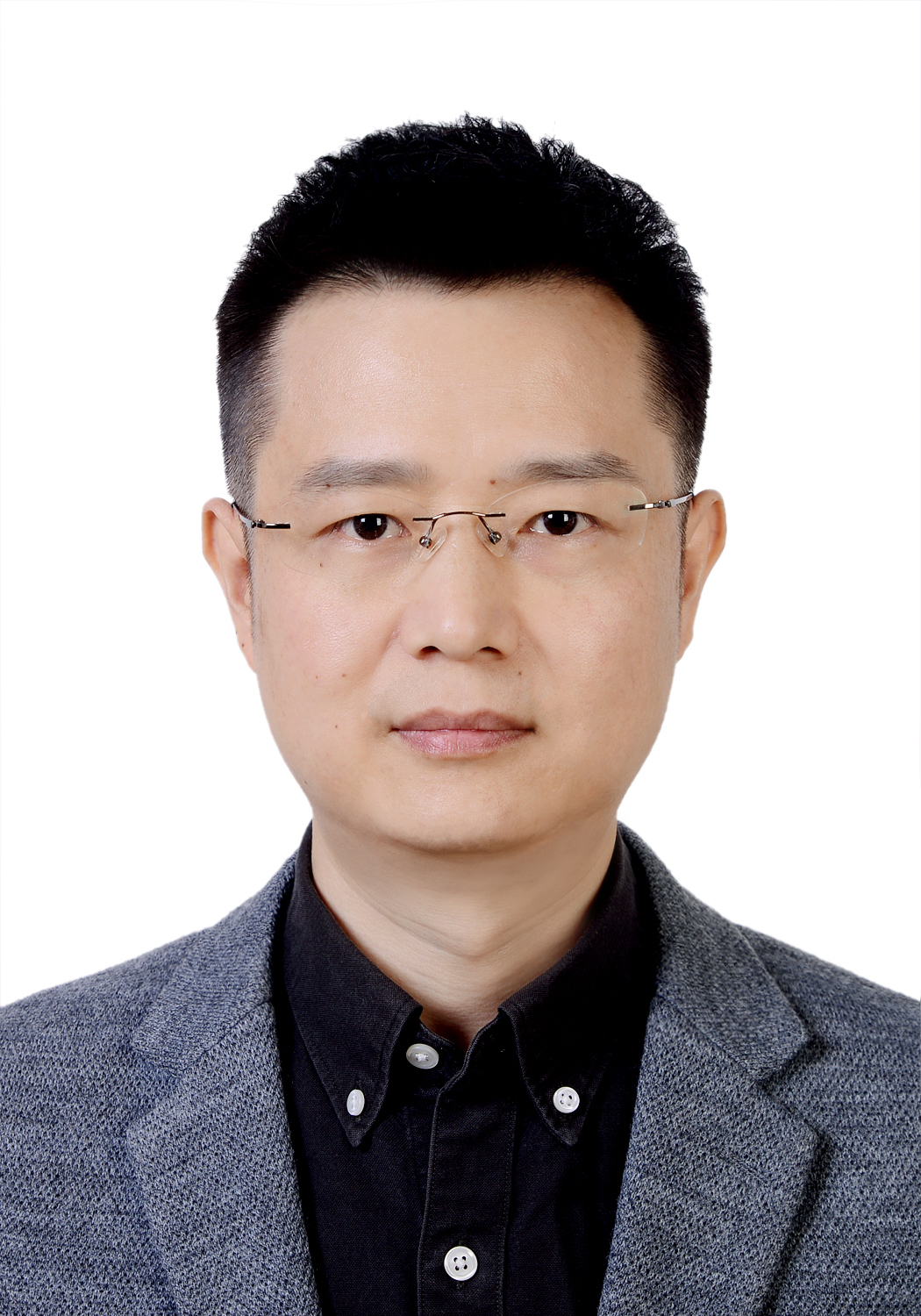}}]
{Zhaoquan Yuan} received his B.E. degree in Computer Science and Technology from the University of Science and Technology of China (USTC), Hefei, China, and his Ph.D. degree in Pattern Recognition and Intelligent Systems from the Institute of Automation, Chinese Academy of Sciences.
Currently, he is an Associate Professor in the School of Computing and Artificial Intelligence at Southwest Jiaotong University, Chengdu, China. 
His research interests include computer vision, multimodal semantic analysis, and deep learning.
\end{IEEEbiography}

\vspace{11pt}
\vspace{-30pt}

\begin{IEEEbiography}
[{\includegraphics[width=1in, height=1.25in, clip, keepaspectratio]{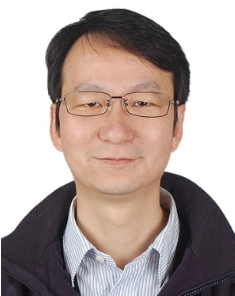}}]
{Xiao Wu}(S\textsuperscript{'}05–M\textsuperscript{'}08)
received the B.Eng. and M.S. degrees in computer science from Yunnan Uni- versity, Yunnan, China, in 1999 and 2002, respectively, and the Ph.D. degree in computer science from the City University of Hong Kong, Hong Kong, in 2008.

He was with the Institute of Software, Chinese Academy of Sciences, Beijing, China, from 2001 to 2002. He was a Research Assistant and a Senior Research Associate with the City University of Hong Kong, Hong Kong, from 2003 to 2004 and
from 2007 to 2009, respectively. He was with the School of Computer Science, Carnegie Mellon University, Pittsburgh, PA, USA, and with the School of Information and Computer Science, University of California at Irvine, Irvine, CA, USA, as a Visiting Scholar, from 2006 to 2007 and 2015 to 2016, respectively. He is currently a Professor and the Assistant Dean with the School of Computing and Artificial Intelligence, Southwest Jiaotong University, Chengdu, China. He has authored or co-authored over 70 research papers in well-respected journals, such as TIP, TMM, TMI, and prestigious proceedings like CVPR and ACM MM. His research interests include multimedia information retrieval, image/video computing, and computer vision. He was a recipient of the Second Prize of Natural Science Award of the Ministry of Education, China, in 2016, and the Second Prize of Science and Technology Progress Award of Henan Province, China, in 2017.
\end{IEEEbiography}

\vspace{11pt}
\vspace{-30pt}

\begin{IEEEbiography}
[{\includegraphics[width=1in,height=1.25in,clip,keepaspectratio]{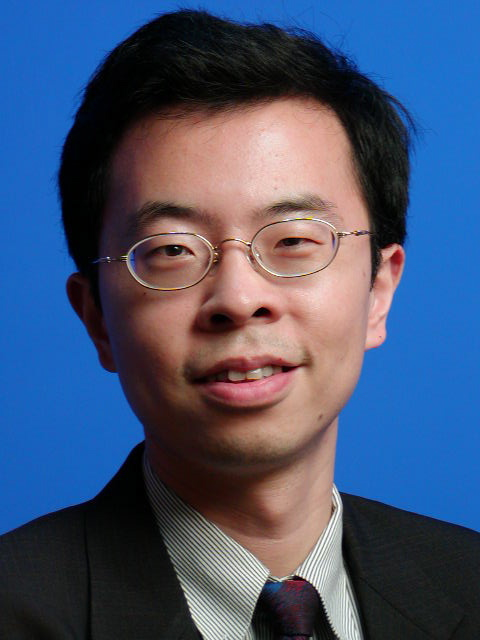}}]{Changsheng Xu}
(M\textsuperscript{'}97 - SM\textsuperscript{'}99 - F\textsuperscript{'}14) is a Professor with the National Lab of Pattern Recognition, Institute of Automation, Chinese Academy of Sciences and the Executive Director of China-Singapore Institute of Digital Media, Singapore. He holds 30 granted/pending patents and has authored or coauthored more than 200 refereed research papers. His research interests include multimedia content anaysis/indexing/retrieval, pattern recognition, and computer vision. 
Prof. Xu is a Fellow of the IAPR and an ACM Distinguished Scientist. He is an Associate Editor of the IEEE TRANSACTIONS ON MULTIMEDIA, ACM Transactions on Multimedia Computing, Communications and Applications, and ACM/Springer Multimedia Systems Journal. He was the recipient of the Best Associate Editor Award of ACM Transactions on Multimedia Computing, Communications and Applications in 2012 and the Best Editorial Member Award of ACM/Springer Multimedia Systems Journal in 2008. He served as the Program Chair of ACM Multimedia 2009. He has served as an Associate Editor, Guest Editor, General Chair, Program Chair, Area/Track Chair, Special Session Organizer, Session Chair, and TPC Member for more than 20 IEEE and ACM multimedia journals, conferences, and workshops.
\end{IEEEbiography}

\vfill

\end{document}